\documentclass[12pt, a4paper, onecolumn]{article}

\usepackage{fullpage}
\usepackage{graphicx}
\usepackage[english]{babel}
\usepackage[center]{caption}
\usepackage{mathtools}
\usepackage{mathptmx}    
\usepackage{latexsym}           
\usepackage{url}
\usepackage{multirow}
\usepackage{amssymb}
\usepackage{amsthm}
\usepackage{amssymb}
\usepackage{amsmath}
\usepackage{lineno}

\usepackage{algorithmic}
\usepackage{algorithm}

\long\def\symbolfootnote[#1]#2{
\begingroup
	\def\thefootnote{\fnsymbol{footnote}}\footnote[#1]{#2}
\endgroup}

\begin{document}

\title{Interference Effects in Quantum Belief Networks}

\author{Catarina Moreira\\ \small \texttt{catarina.p.moreira@ist.utl.pt}\\
\and
Andreas Wichert\\ \small \texttt{andreas.wichert@ist.utl.pt}
\and
\\Instituto Superior T\'{e}cnico, INESC-ID\\ Av. Professor Cavaco Silva, 2744-016 Porto Salvo, Portugal\\  
\\ \small The original publication is available at: Applied Soft Computing, Elsevier\\
 \small \text{\url{http://www.sciencedirect.com/science/article/pii/S095741741300238}}
}

\date{}

\maketitle

\begin{abstract}
Probabilistic graphical models such as Bayesian Networks are one of the most powerful structures known by the Computer Science community for deriving probabilistic inferences. However, modern cognitive psychology has revealed that human decisions could not follow the rules of classical probability theory, because humans cannot process large amounts of data in order to make judgements. Consequently, the inferences performed are based on limited data coupled with several heuristics, leading to violations of the law of total probability. This means that probabilistic graphical models based on classical probability theory are too limited to fully simulate and explain various aspects of human decision making. 

Quantum probability theory was developed in order to accommodate the paradoxical findings that the classical theory could not explain. Recent findings in cognitive psychology revealed that quantum probability can fully describe human decisions in an elegant framework. Their findings suggest that, before taking a decision, human thoughts are seen as superposed waves that can interfere with each other, influencing the final decision.
		
In this work, we propose a new Bayesian Network based on the psychological findings of cognitive scientists. We made experiments with two very well known Bayesian Networks from the literature. The results obtained revealed that the quantum like Bayesian Network can affect drastically the probabilistic inferences, specially when the levels of uncertainty of the network are very high (no pieces of evidence observed). When the levels of uncertainty are very low, then the proposed quantum like network collapses to its classical counterpart.
\end{abstract}

\symbolfootnote[0]{This work was supported by national funds through FCT - Funda\c{c}\~{a}o para a Ci\^{e}ncia e a Tecnologia, under project PEst-OE/EEI/LA0021/2013}

\section{Introduction}\label{sec:intro}

The problem of violations of the axioms of probability go back to the early 60’s.~\cite{Ellsberg61} published a work that influenced modern psychology by showing that humans violate the laws of probability theory when making decisions under risk. The principle that humans were constantly violating is defined by \emph{The Sure Thing Principle}. It is a concept widely used in game theory and was originally introduced by~\cite{savage54}. This principle is fundamental in Bayesian probability theory and states that if one prefers action $A$ over $B$ under state of the world $X$, and if one also prefers $A$ over $B$ under the complementary state of the world $X$, then one should always prefer action $A$ over $B$ even when the state of the world is unspecified.

Cognitive psychologists A. Tversky and D. Khamenman also explored more situations where classical probability theory could not be accommodated in human decisions. In their pioneering work,~\cite{Tversky74} realised that the beliefs expressed by humans could not follow the rules of Boolean logic or classical probability theory, because humans cannot process large amounts of data in order to make estimations or judgements. Consequently, the inferences performed are based on limited data coupled with several heuristics, leading to a violation on one of the most important laws in bayesian theory: the law of total probability.

\begin{figure}[h!]
\resizebox{\columnwidth}{!} {
\includegraphics{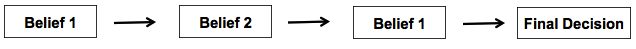}
}
\caption{Example of how information is processed in a classical setting. At each time, beliefs can only be in one definite state.}
\label{fig:definite_states}
\end{figure}

One of the key differences between classical and quantum theories is the way how information is processed. According to classical decision making, a person changes beliefs at each moment in time, but it can only be in one precise state with respect to some judgement. So, at each moment, a person is favouring a specific belief. The process of human inference deterministically either jumps between definite states or stays in a single definite state across time~\cite{Busemeyer12book}. Most computer science, cognitive and decision systems are modelled according to this single path trajectory principle. Figure~\ref{fig:definite_states} illustrates this idea.
 
\begin{figure}[h!]
\resizebox{\columnwidth}{!} {
	\includegraphics{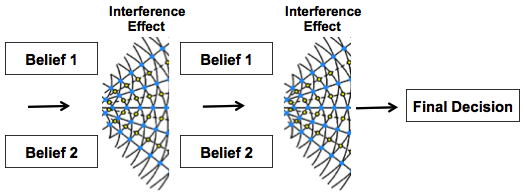}
}
\caption{In human decision making, believes occur in the human mind at the same time, leading to uncertainty feelings and ambiguity. Beliefs can be represented in superposition states that can generate interferences between them.}
\label{fig:interference}
\end{figure}

In quantum information processing, on the other hand, information (and consequently beliefs) are modelled via wave functions and therefore they cannot be in definite states. Instead, they are in an indefinite quantum state called the $superposition$ state. That is, all beliefs are occurring on the human mind at the same time. According to cognitive scientists, this effect is responsible for making people experience uncertainties, ambiguities or even confusion before making a decision.  At each moment, one belief can be more favoured than another, but all beliefs are available at the same time. In this sense, quantum theory enables the modelling of the cognitive system as it was a wave moving across time over a state space until a final decision is made. From this superposed state, uncertainty can produce different waves coming from opposite directions that can crash into each other, causing an interference distribution. This phenomena can never be obtained in a classical setting. Figure~\ref{fig:interference} exemplifies this. When the final decision is made, then there is no more uncertainty. The wave collapses into a definite state. Thus, quantum information processing deals with both definite and indefinite states~\cite{Busemeyer12book}.

\subsection{Motivation: Violations in The Two-Stage Gamblings}

\cite{Tversky92} were one of the first researchers to test the veracity of Savage's principle under human cognition in a gambling game. In their experiment, participants were asked at each stage to make the decision of whether or not to play a gamble that has an equal chance of winning \$200 or losing \$100. Figure~3 illustrates the experiment. Three conditions were verified:

\begin{enumerate}
	\item Participants were informed if they had won the first gamble;
	\item Participants were informed if they had lost the first gamble;
	\item Participants did not know the outcome of the first gamble;
\end{enumerate}

\begin{figure}[h!]
	\centering
	\includegraphics[scale=0.4]{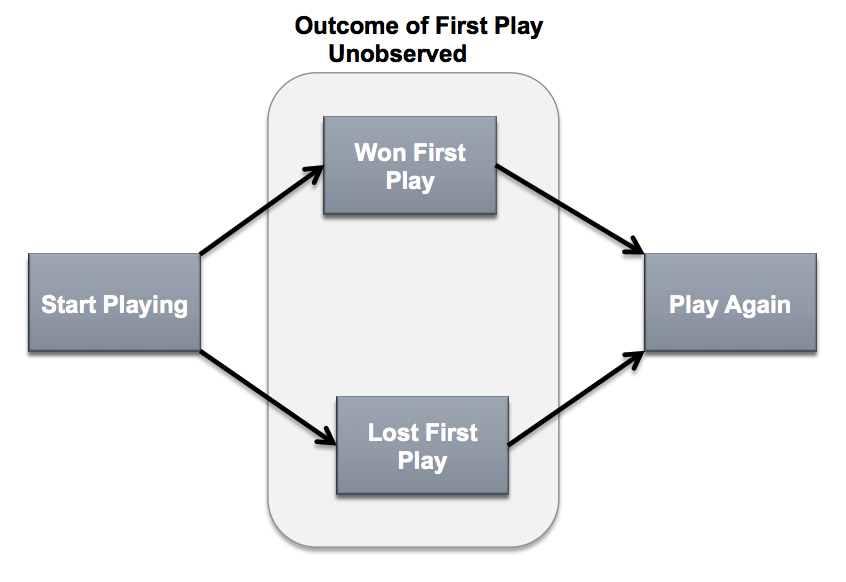}
	\label{fig:two_stage_gambling}
	\caption{The two-stage gambling experiment proposed by~\cite{Tversky92}}
\end{figure}

The two-stage gambling game was one of the first experiments used in order to determine if the Sure Thing Principle would be verified even with people that did not know about the existence of this principle. The results obtained  in~\cite{Tversky92} experiment showed that this principle is constantly being violated and consequently humans do not perform inferences according to the laws of probability theory and Boolean logic. 

The overall results revealed that participants who knew that they won the first gamble, decided to play again. Participants who knew that they lost the first gamble, also decided to play again. Through Savage's sure thing principle, it was expected that the participants would choose to play again, even if they did not know the outcome of the first gamble. However, the results obtained revealed something different. If the participants did not know the outcome of the first gamble, then many of them decided not to play the second one. 

Several researchers replicated this experiment. The overall results are specified in Table~\ref{tab:game_results}.

\begin{table}[h!]
	\resizebox{\columnwidth}{!} {
		\begin{tabular}{l | c | c | c }

			\textbf{Literature}	& \textbf{Pr(Play Again $|$ Won 1st Play)}	& \textbf{Pr(Play Again $|$ Lost 1st Play)}	& \textbf{Pr(Play Again $|$ Unknown 1st Play)} \\
			
			\hline
			
			\cite{Tversky92}			& 69\%					& 58\%   				& 37\%    \\
			\cite{Kuhberger01}			& 72\% 					& 47\%				& 48\%	\\
			\cite{Lambdin07}			& 63\%					& 45\%				& 41\%	\\
			\hline
			\textbf{Average}			& 68\%					& 50\%				& 42\%	\\
			\hline

		\end{tabular}
	}
	\caption{Observations reported by different works in the literature about the two-step gambling game.}
	\label{tab:game_results}
\end{table}

Why did the findings reported in Table~\ref{tab:game_results}  generate so much controversy in the scientific community? Because, the data observed is not in accordance with the classical law of total probability. In Tversky and Shafir's experiment~\cite{Tversky92}, the probability of a participant playing the second gamble, given that the outcome of the first gamble is unknown, $Pr(G | U)$, can be computed through the law of total probability:  

\begin{equation}
Pr(~G~|~U~) = Pr(~W~|~U~) \cdot Pr(~G~|~W~) + Pr(~L~|~U~) \cdot Pr(~G~|~L~)
\label{eq:law_total_prob}
\end{equation}

In Equation~\ref{eq:law_total_prob}, $Pr(W | U)$ corresponds to the probability of a player winning the first gamble, given that (s)he participated on the game in the first place. $Pr(G | W)$ is the probability of playing the second gamble, given that it is known that the player won the first one. $Pr(L | U)$ corresponds to the probability of losing the first gamble, given that the participant decided to play the game in the first place. And finally, $Pr(G | L)$ is the probability of a participant playing the second gamble, given that it is known that (s)he lost the first one. 

Following the law of total probability in Equation~\ref{eq:law_total_prob}, the probability of playing the second gamble, given that the player did not know the outcome of the first one, should be between the following values~\cite{Busemeyer12book}:

\begin{equation}
Pr(~G~|~W~) \ge Pr(~G~|~U~) \ge Pr(~G~|~L~) 
\label{eq:relation_classic}
\end{equation}

The findings reported by~\cite{Tversky92}, however, revealed a different relation. Equation~\ref{eq:relation_observed} demonstrates that this relation is violating one of the most fundamental laws of Bayesian probability theory:

\begin{equation}
Pr(~G~|~W ) = 0.69 \ge Pr(~G~|~L~) = 0.58 \ge Pr(~G~|~U~) = 0.37
\label{eq:relation_observed}
\end{equation}

\cite{Tversky92} explained these findings in the following way: when the participants knew that they won, then they had extra house money to play with and decided to play the second round. If the participants knew that they lost, then they chose to play again with the hope of recovering the lost money. But, when the participants did not know if they had won or lost the first gamble, then these thoughts, for some reason, did not emerge in their minds and consequently they decided not to play the second gamble. Other works in the literature also replicated this two-stage gambling experiment~\cite{Shafir92,Kuhberger01,Lambdin07}, also reporting similar results to~\cite{Tversky92}. Their results are summarised in Table~\ref{tab:game_results}. 

There have been different works in the literature trying to explain and model this phenomena~\cite{Busemeyer12book, Busemeyer09,Busemeyer09markov}. Although the models in the literature diverge, they all agree in one thing: one cannot use classical probability theory to model this phenomena, since the most important rules are being violated. This two stage gambling game experiment was one of the most important works that motivated the use of different theories outside of classical bayesian theory and boolean logic, more specifically the usage of quantum probability theory.

\subsection{Research Questions}\label{sec:rq}

Recent findings in the cognitive psychology literature revealed that humans are constantly violating the law of total probability when making decisions under risk~\cite{Asano12,Busemeyer09markov,busemeyer06}. These researchers also showed that quantum probability theory enables the development of decision models that are able to simulate human decisions. Given that most of the systems that are used nowadays are based on Bayesian probability theory, is it possible to achieve better inference mechanisms in these systems using quantum probability theory? For instance, many medical diagnosing systems are based in classical probabilistic graphical models such as Bayesian Networks. Can one achieve better performances in diagnosing patients using quantum probability?

Generally speaking, a Bayesian Network is a probabilistic graphical model that represents a set of random variables and their conditional dependencies via a directed acyclic graph.

There are two main works in the literature that have contributed to the development and understanding of Quantum Bayesian Networks. One belongs to~\cite{Tucci95} and the other to~\cite{Leifer08}. 

In the work of~\cite{Tucci95}, it is argued that any classical Bayesian Network can be extended to a quantum one by replacing real probabilities with quantum complex amplitudes. This means that the factorisation should be performed in the same way as in a classical Bayesian Network.  One big problem with Tucci's work is concerned with the inexistence of any methods to set the phase parameters. The author states that, one could have infinite Quantum Bayesian Networks representing the same classical Bayesian Network depending on the values that one chooses to set the parameters. This requires that one knows \emph{a priori} which parameters would lead to the desired solution for each node queried in the network (which we never know).

In the work of~\cite{Leifer08}, the authors argue that, in order to develop a quantum Bayesian Network, it is required a quantum version of probability distributions, quantum marginal probabilities and quantum conditional probabilities. The proposed model fails to provide any advantage relatively to the classical models, because it cannot take into account interference effects between unobserved random variables. In the end, both models provide no advantages in modelling decision making problems that try to predict decisions that violate the laws of total probability.

\begin{figure}
	\centering
	\includegraphics[scale=0.6]{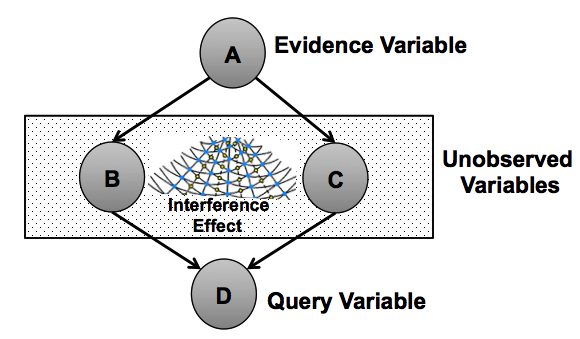}
	\label{fig:qbn_example1}
\caption{An example of the proposed model. In a quantum Bayesian Network we will asusume that if some nodes are unobserved, then the inference process is propagated like a wave through both nodes like a superposed state. Interference effects my arise since the waves can interfere with each other.}
\end{figure}

In this paper, the core of the proposed Bayesian Network is based on the psychological findings uncovered in the works of~\cite{Busemeyer09markov,busemeyer06,Busemeyer12book, Busemeyer09} and on quantum information processing. These authors show that, before taking a decision, human thoughts are seen as superposed waves that can interfere with each other, influencing the final decision. In Bayesian Networks, nodes can either be query variables, evidences or simply unknown. Given that we do not observe the unknown nodes of a Bayesian Network, since we do not know for sure what values they can take, then what would happen to the inference process if these nodes are put in a representation of a quantum superposition and interfere with each other (Figure~4)? Can a better inference process be achieved? These are the main research questions that this paper aims at answering. So far, to the best of our knowledge, there is no previous work in the Computer Science community that attempts to map these psychological findings into computer science decision making systems, such as Bayesian Networks.

In order to validate our hypothesis, we performed experiments with well known classical Bayesian Networks from the literature. We first create a quantum Bayesian Network that can accommodate the paradoxical findings in the two-stage gambling game. We then generalise our quantum Bayesian Network in order to deal with larger and more complex datasets that are used in the literature: the well known Burglar/Alarm Bayesian Network from~\cite{russel10} and the Lung Cancer Bayesian Network from~\cite{Pearl88}.

\subsection{Outline}

Before describing the proposed model, we first need to introduce some quantum probability concepts for the understanding of this work. Sections~\ref{sec:probability} and~\ref{sec:traj} present the main differences between classical and quantum probability theory. Instead of just presenting a set of formulas, we show this difference by means of an illustrative example, just like proposed in~\cite{Busemeyer12book}. In Section~\ref{sec:interference}, we describe how beliefs can act like waves and interfere with each other. We show mathematically how this interference term can be derived by using well known rules of complex numbers. Section~\ref{sec:theta_lit} addresses the main works of the literature that contributed for the development of the interference term. It also introduces a new interference formula that will be applied in the proposed quantum probabilistic graphical models. Section~\ref{sec:classic_vs_quantum_gamble} presents a comparison between a classical Bayesian Network model against the proposed quantum interference Bayesian Network applied to the problem of two-stage gambles. Section~\ref{sec:classic_vs_quantum_burglar} presents another comparison between the classical and quantum Bayesian Networks, but for a more complex network from the literature. In Section~\ref{sec:discussion}, it is made a discussion about the results obtained in the experiments performed in Section~\ref{sec:classic_vs_quantum_burglar}. Section~\ref{sec:opt} presents an additional experiment over another Bayesian Network in order to study the impact of the quantum interference parameters in different scenarios. Section~\ref{sec:rel_work} presents the most relevant works of the literature. Finally, Section~\ref{sec:conclusions} presents the main conclusions of this work.

\section{Probability Axioms of Classical and Quantum Theory}\label{sec:probability}

In this section, we describe the main differences between classical theory and quantum probability theory through examples. The example analyzed concerns jury duty. Suppose you are a juror and you must decide whether a defendant is guilty or innocent. The following sections describe how the classical and quantum theory evolve in the inference process. All this analysis is based on the book of~\cite{Busemeyer12book}.

\subsection{Space}

\begin{figure}[h!]
	\parbox{.48\linewidth}{
	\centering
	\includegraphics[scale=0.45]{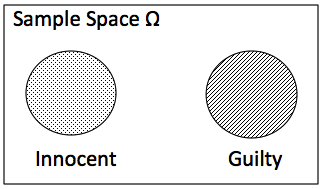}
	\caption{In classical probability theory, results are contained in sample spaces.}
	\label{fig:sample_space}
	}
	\hfill
	\parbox{.48\linewidth}{
	\centering
	\includegraphics[scale=0.4]{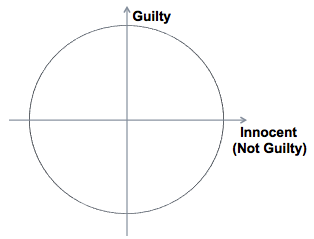}
	\caption{In quantum probability theory, events are spanned by a set of orthornormal basis vectors in a Hilbert Space.}
	\label{fig:hilbert_space}	
	}
\end{figure}

In classical probability theory, events are contained in Sample Spaces.
A Sample Space $\Omega$ corresponds to the set of all possible outcomes of an experiment or random trial~\cite{DeGroot11Prob}. For example,  when judging whether a defendant is guilty or innocent, the sample space is given by $\Omega = \{Guilty, Innocent\}$. Figure~\ref{fig:sample_space} presents a diagram showing the sample space of a defendant being guilty or innocent.\\

In quantum probability theory, events are contained in the so called Hilbert Spaces. 
A Hilbert Space $H$ can be viewed as a generalisation and extension of the Euclidean space into spaces with any finite or infinite number or dimensions. It can be see as a vector space of complex numbers and offers the structure of an inner product to enable the measurement of angles and lengths~\cite{mika03}. The space is spanned by a set of orthonormal basis vectors $H = \{ Guilty , Innocent \}$. Together, these vectors form a basis for the space. Figure~\ref{fig:hilbert_space} presents a diagram showing the Hilbert space of a defendant being guilty or innocent~\cite{Busemeyer12book}. Since a Hilbert space enables the usage of complex numbers, then, in order to represent the events $Guilty$ and $Innocent$, one would need  two dimensions for each event (one for the real part and another for the imaginary part). In quantum theory, one usually ignores the imaginary component in order to be able to visualise geometrically all vectors in a 2-dimensional space.

\subsection{Events}

In classical probability theory, events can be defined by a set of outcomes to which a probability is assigned. They correspond to a subset of the sample space $\Omega$ from which they are contained in. Events can be mutually exclusive and they obey to set theory. This means that operations such as intersection or union of events are well defined. Since they respect set theory, the distributive axiom is also defined between sets. In our example, $Guilty$ or $Innocent$ can be seen as two mutually exclusive events.

\begin{figure}[h!]
	\centering
	\includegraphics[scale=0.5]{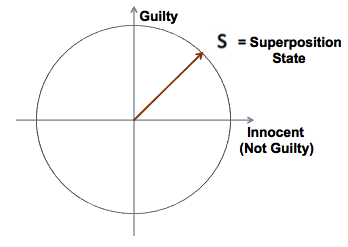}
	\caption{Example of an event represented by a superposition of the states \textit{Guilty} and \textit{Innocent} denoted by the quantum state $S$.}
	\label{fig:quantum_events}
\end{figure}

According to quantum probability theory, events correspond to a subspace spanned by a subset of the basis vectors contained in the Hilbert Space. Events can be orthogonal, that is, they can be mutually exclusive. Operations such as intersection and union of events are well defined if the events are spanned by the same basis vector~\cite{Busemeyer12book}. In quantum theory, all events contained in a Hilbert Space are defined through a superposition state which is represented by a state vector $S$ comprising the occurrence of all events. In our example, $Guilty$ and $Innocent$ correspond to column vectors representing the main axis of the circle in Figure~\ref{fig:quantum_events}. They are defined as follows:
\[  Guilty = \left[ \begin{matrix} ~1~\\ ~0~ \end{matrix} \right] ~~~~~~~~~~~~~~~~~ Innocent = \left[ \begin{matrix} ~0~\\ ~1~ \end{matrix} \right] \]

In Figure~\ref{fig:quantum_events}, the superposition state $S$ can be defined as follows.
\begin{equation}
S = \frac{e^{i\theta_G}}{\sqrt{2}} ~ Guilty + \frac{e^{i\theta_I}}{\sqrt{2}} ~ Innocent 
\label{eq:ampl}
\end{equation}

In Equation~\ref{eq:ampl}, one might be wondering what the $\frac{e^{i\theta}}{\sqrt{2}}$ values mean. They are called $probability$ $amplitudes$. They correspond to the amplitudes of a wave and are described by complex numbers. The $e^{i\theta}$ term is defined as the phase of the amplitude. It can be seen as a shift of the wave. These amplitudes are related to classical probability by taking the squared magnitude of these amplitudes. This is achieved by multiplying the amplitude with its complex conjugate (represented by the symbol $*$).

\begin{equation}
Pr(Guilty) = \left| \frac{e^{i\theta_G}}{\sqrt{2}} \right|^2 = \left(\frac{e^{i\theta_G}}{\sqrt{2}} \right)  \cdot  \left(\frac{e^{i\theta_G}}{\sqrt{2}} \right)^{*}  =\frac{e^{i\theta_G}}{\sqrt{2}}~\cdot~\frac{e^{-i\theta_{G}}}{\sqrt{2}}  = e^{i(\theta_G - \theta_G) } \left( \frac{1}{\sqrt{2}} \right)^2 = 0.5
\end{equation}

 In quantum theory, it is required that the sum of the squared magnitudes of each amplitude equals $1$. This axiom is called the normalization axiom and corresponds to the classical theory constraint that the probability of all events in a sample space should sum to one.
\begin{equation}
\left| \frac{e^{i\theta_G}}{\sqrt{2}} \right|^2 + \left| \frac{e^{i\theta_I}}{\sqrt{2}} \right|^2  = 1
\end{equation}

\subsection{System State}~\label{sec:system_state}

A system state is nothing more than a probability function $Pr$ which maps events into probability numbers, i.e., positive real numbers between $0$ and $1$.

In classical theory, the system state corresponds to exactly its definition. There is a function that is responsible to assign a probability value to the outcome of an event. If the event corresponds to the sample space, then the system state assigns a probability value of $1$ to the event. If the event is empty, then it assigns a probability of $0$. In our example, if nothing else is told to the juror, then the probability of the defendant being guilty is $Pr(Guilty) = 0.5$.

\begin{figure}[ht!]
	\centering
	\includegraphics[scale=0.45]{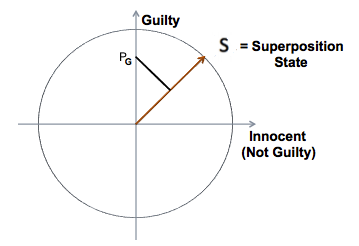}
	\caption{Geometric representation of a projection into the \textit{Guilty} subspace. The computation of the probability starts in the superposition state $S$. Then,  this superposition state is projected into the $Guily$ subspace $P_G$. The final probability corresponds to the squared magnitude of the projection.}
	\label{fig:hilbert_system_state}
\end{figure}

In quantum theory, the probability of a defendant being $Guilty$ is given by the squared magnitude of the projection from the superposition state $S$ to the subspace containing the observed event $Guilty$. Figure~\ref{fig:hilbert_system_state} shows an example. If nothing is told to the juror about the guiltiness of a defendant, then according to quantum theory, we start with a superposition state $S$.

\[ S  = \frac{e^{i\theta_G}}{\sqrt{2}} ~ Guilty + \frac{e^{i\theta_I}}{\sqrt{2}} ~ Innocent \]

When someone asks whether the defendant is guilty, then we project the superposition state $S$ into the relevant subspace, in this case the $Guilty$ subspace $(P_G)$, just like shown in Figure~\ref{fig:hilbert_system_state}. The probability is simply the squared magnitude of the projection, that is:

\[ Pr(Guilty) = |P_G|^2 = \left| \frac{e^{i\theta_G}}{\sqrt{2}} \right|^2 = 0.5  \]

Which has exactly the same outcome as in the classical theory.

\subsection{State Revision}

State revision corresponds to the situation where after observing an event, we are interested in observing other events given that the previous one has occurred.

In classical theory, this is addressed through the conditional probability formula $Pr(B|A)=\frac{Pr(A \cap B)}{Pr(B)}$. So, returning to our example, suppose that some evidence has been given to the juror proving that the defendant is actually guilty, then what is the probability of him being innocent? This is computed in the following way.

\[ Pr(Innocent | Guilty) = \frac{Pr(Innocent \cap Guilty)}{Pr(Guilty)} = 0 \]

Since the events $Guilty$ and $Innocent$ are mutually exclusive, then their intersection is empty, leading to a zero probability value.

In quantum theory, the state revision is given by first projecting the superposition state $S$ into the subspace representing the observed event. Then, the projection is normalised such that the resulting vector is unit length. Again, if we want to determine the probability of a defendant being innocent, given he was found guilty, the calculations are performed as follows. We first start in the superposition state vector $S$.

\[ S = \frac{e^{i\theta_G}}{\sqrt{2}}  ~Guilty + \frac{e^{i\theta_I}}{\sqrt{2}}  ~ Innocent \]
	
Then, we observe that the defendant is guilty, so we project the state vector $S$ into the $Guilty$ subspace and normalise the resulting projection.
	
\[ S_G = \frac{(e^{i\theta_G}/\sqrt{2}) ~ Guilty }{\sqrt{\left| e^{i\theta_G} / \sqrt{2} \right|^2}} \]

\[ S_G = e^{i\theta_G} ~ Guilty  + 0 ~ Innocent \]

From the resulting state, we just extract the probability of being innocent by simply squaring the respective probability amplitude. Again, we obtain the same results as the classical theory.

\[ Pr(Innocent|Guilty) = |~0~|^2 = 0 \]

\begin{figure}[ht!]
	\centering
	\includegraphics[scale=0.45]{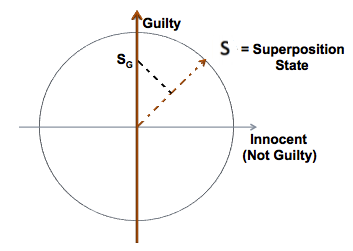}
	\caption{Geometric representation of a state revision. Note that the new state is represented uniquely by the $Guilty$ subspace.}
	\label{fig:hilbert_state_revision}
\end{figure}

\section{The Path Trajectory Principle}\label{sec:traj}

In order to describe direct dependencies between a set of variables, path diagrams are generally used. This section shows how to compute quantum probabilities in a Markov model using Feynman's path rules, just like presented in the work of~\cite{Busemeyer09markov}. 

\subsection{Single Trajectories}

\begin{figure}[ht]
	\parbox{.3\linewidth}{
	\centering
	\includegraphics[scale=0.45]{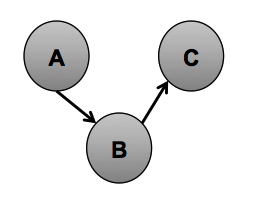}
	\caption{Single path trajectory}
	\label{fig:single_trajectory}
	}
	\hfill
	\parbox{.35\linewidth}{
	\centering
	\includegraphics[scale=0.45]{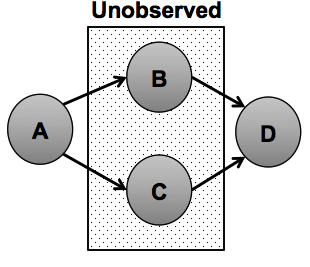}
	\caption{Multiple indistinguishable paths}
	\label{fig:multiple_trajectory_unobs}	
	}
	\hfill
	\parbox{.3\linewidth}{
	\centering
	\includegraphics[scale=0.45]{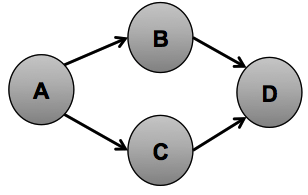}
	\caption{Multiple distinguishable paths}
	\label{fig:multiple_trajectory_obs}	
	}
\end{figure}

Consider the diagram represented in Figure~\ref{fig:single_trajectory}. 

The computation of the probability of transiting from an initial state $A$ to a final state $C$, transiting from an intermediate state $B$, can be achieved through a classical Markov model. The probability can be computed by making the product of the individual probabilities for each transition, from one state to another, through the usage of conditional probabilities. According to Figure~\ref{fig:single_trajectory}, the probability of transiting from state $A$, followed by state $B$ and ending in state $C$, that is, $Pr(A \rightarrow B \rightarrow C)$, is given by:

\begin{equation}
Pr(~A \rightarrow B \rightarrow C~) = Pr(~A~) \cdot Pr( ~B~| ~A~ ) \cdot Pr( ~C~|~B~ ) 
\end{equation}

In a quantum path diagram model, the computation of the probabilities for a single path trajectory is similar to the classical Markov model. The calculation can be performed using Feynman's first rule, which asserts that the probability of a single path trajectory consists in the product of the squared magnitudes of the amplitudes for each transition from one state to the next along the path. This means that the quantum probability value of a single path trajectory is the same as the classical Markov probability for the same path. In Equation~\ref{eq:feynman1} and through the rest of this work, complex probability amplitudes will be represented by the symbol $\psi$.

\begin{equation}
Pr(~A \rightarrow B \rightarrow C~) =  |  ~\psi_{A}  |^2 \cdot |~\psi_{~B~| ~A~ }|^2  \cdot | ~\psi_{ ~C~|~B~ }|^2 = Pr( A )\cdot Pr( ~B~| ~A~) \cdot Pr( ~C~|~B~ ) 
\label{eq:feynman1}
\end{equation}

\subsection{Multiple Indistinguishable Trajectories}

An indistinguishable path consists in transiting from an initial state $A$ to a final state $D$ by transiting from multiple possible paths without knowing for certain which path was taken to reach the goal state. Figure~\ref{fig:multiple_trajectory_unobs} shows an example of multiple indistinguishable trajectories. In a classical Markov model, if one does not observe which path was taken to reach the final state $D$, then one simply computes this probability by summing the individual probabilities of each path. This is known as the single path principle and is in accordance with the law of total probability. So, following Figure~\ref{fig:multiple_trajectory_unobs}, in order to reach state $D$ starting in state $A$, one can take the path $ A \rightarrow B \rightarrow D $ or the path $A \rightarrow C \rightarrow D$. The final probability is given by:

\begin{equation}
 Pr(~ A \rightarrow D~) = Pr(~A~) \cdot Pr( ~B~| ~A~ ) \cdot Pr( ~D~|~B~ ) +  Pr(~A~) \cdot Pr( ~C~| ~A~ ) \cdot Pr( ~D~|~C~ ) 
\end{equation}

Quantum probability theory rejects the single path trajectory principle. If one does not observe which path was taken to reach the goal state, then one cannot assume that one of each possible paths was used. Instead, quantum probability argues that, when the path is unobserved, then the goal state can be reached through a superposition of path trajectories. This is known as Feynman's second rule, which states that the amplitude of transiting from an initial state $A$ to a final state $D$, taking multiple indistinguishable paths, is given by the sum of all amplitudes for each path. This rule is in accordance with the law of total amplitude and the probability is computed by taking the squared magnitude of this sum. This probability is not equal to the classical Markov model.
\[Pr(~ A \rightarrow D~) = |\psi_A \cdot \psi_{B~| ~A} \cdot  \psi_{D~|~B} + \psi_A \cdot \psi_{ ~C~| ~A}~ \cdot ~ \psi_{ ~D~|~C} ~|^2 =  \]
\begin{equation}
 = |~\psi_A \cdot \psi_{B~| ~A} \cdot \psi_{~D~|~B~}  ~|^2 + |~\psi_A \cdot \psi_{C~| ~A}~ \cdot ~ \psi_{ D~|~C} ~|^2 + 2~ \cdot |~\psi_A \cdot \psi_{B~| ~A}~ \cdot ~ \psi_{D~|~B}~|~ \cdot ~|~\psi_A \cdot \psi_{C~| ~A} ~ \cdot ~ \psi_{ D ~ |~C~ }~|\cos(\theta)
\end{equation}

The term $\cos(\theta)$ is the inner product between the vectors formed by $ |~\psi_{B~| ~A}~ \cdot ~\psi_{~D~|~B~}  ~|$ and $|~\psi_{C~| ~A}~ \cdot ~ \psi_{ D~|~C} ~|$. It comes from Euler's rule: $\cos( \theta ) = \left( e^{i\theta} + e^{-i\theta} \right) / 2$ and corresponds to a quantum interference term that does not exist in classical probability theory. Section~\ref{sec:interference} details how this term is derived. Since the interference term can lead to estimations with values higher than 1, then it is necessary to normalize this value in order to obtain a probability value. 

When the path is observed, quantum probability theory collapses to the classical Markov model. This is know as Feynman's third rule and states that the probability amplitude of observed multiple path trajectories corresponds to the sum of the amplitudes of each individual path. The probabilities are then taken by making the squared magnitude of each individual path. Figure~\ref{fig:multiple_trajectory_obs} illustrates this example:
 
\[Pr(~ A \rightarrow D~) = |~\psi_A \cdot  \psi_{ B~| ~A} ~ \cdot ~\psi_{ ~D~|~B~ } ~|^2~+  ~|~\psi_A \cdot \psi_{ ~C~| ~A~ }~ \cdot ~\psi_{ ~D~|~C~ }~|^2  = \]
\begin{equation}
 = Pr( A ) \cdot Pr( ~B~| ~A~ ) \cdot ~Pr( ~D~|~B~ ) +  Pr( A ) \cdot Pr( ~C~| ~A~ ) \cdot Pr( ~D~|~C~ )
\end{equation}

\section{The Interference Term}\label{sec:interference}

Quantum theory enables the modeling of the decision system as a wave moving across time over a state space until a final decision is made. Under this perspective, interference can be regarded as a chain of waves in a superposition state, coming from different directions. When these waves crash, one can experience a destructive effect (one wave destroys the other) or a constructive effect (one wave merges with another). In either case, the final probabilities of each wave is affected. Psychological findings showed that this not only occurs in a microscopic scale (such as electrons), but also occurs at a macroscopic setting~\cite{Khrennikov06,Busemeyer09,busemeyer06}.

In this section we show how to derive the interference term by just taking into account well known properties of complex numbers~\cite{Khrennikov09sure}. The interference term can be derived in two different ways: (1) from the general probability formula of the union of $N$ mutually exclusive events and (2) through the total law of probability. 

\subsection{Deriving the Interference Term from the Union of $N$ Mutually Exclusive Events}

For simplicity, we will start by deriving the interference term for 2 events and then we will generalize for $N$ events. The classical probability formula of the union of two mutual exclusive events is given by:

\begin{equation}
Pr( A \cup B) = Pr(A) + Pr(B) 
\label{eq:union}
\end{equation}

And the relation between a classical probability density function and a quantum probability amplitude is given by Born's rule, that is:

\begin{equation} 
Pr( A  )   = |~e^{i\theta_A}\psi_{A}~|^2 
\label{eq:relation}
\end{equation}

Again, $| e^{i\theta_A}\psi_A |^2$ corresponds to the square magnitude of a complex amplitude. It is obtained by multiplying the probability amplitude with its complex conjugate. That is, $ |~e^{i\theta_A}\psi_{A}~|^2 = e^{i\theta_A}\psi_{A}e^{-i\theta_A}\psi_{A} $.  

Taking into account Equation~\ref{eq:relation}, one can write a superposition state between two mutual exclusive events $A$ and $B$ in the following way:

\begin{equation}
\psi_{A+B} = e^{i\theta_A}\psi_A + e^{i\theta_B}\psi_B 
\label{eq:qn_union}
\end{equation}

The relation of this superposed state with classical probability theory remains:

\begin{equation}
Pr(A \cup B ) = Pr(~A~) + Pr(~B~) \propto |~ e^{i\theta_A}\psi_A + e^{i\theta_B}\psi_B ~| ^2 = \left| \psi_{A+B} \right| ^2
\label{eq:relation_2}
\end{equation}

The quantum counterpart of the classical probability of the union of two mutually exclusive events, when we do not observe them, collapses to Feynmann's second rule, that is:

\[ Pr(A \cup B ) = |~e^{i\theta_A}\psi_{A} + e^{i\theta_B}\psi_{B} ~| ^2 = ( ~e^{i\theta_A}\psi_{A} + e^{i\theta_B}\psi_{B} ~) \cdot (~e^{i\theta_A}\psi_{A} + e^{i\theta_B}\psi_{B} ~)^{*} \]

\[ |~e^{i\theta_A}\psi_{A} + e^{i\theta_B}\psi_{B} ~| ^2 = e^{i\theta_A}\psi_{A}^{~}~ \cdot ~e^{-i\theta_A}\psi_{A} + e^{i\theta_A}\psi_{A}~\cdot~e^{-i\theta_B}\psi_{B} + e^{i\theta_B}\psi_{B}~\cdot~e^{-i\theta_A}\psi_{A} +  e^{i\theta_B}\psi_{B}^{~}~\cdot~e^{-i\theta_B}\psi_{B}  \]

\begin{equation}
 |~e^{i\theta_A}\psi_{A} +e^{i\theta_B} \psi_{B} ~| ^2 = |~\psi_{A}~|^2 + |~\psi_{B}~|^2 + e^{i\theta_A}\psi_{A}~\cdot~e^{-i\theta_B}\psi_{B} + e^{i\theta_B}\psi_{B}~\cdot~e^{-i\theta_A}\psi_{A}
\label{eq:int_1}
\end{equation}

The classical probability of the union of two mutual exclusive events is $P( A \cup B) = Pr(A) + Pr(B)$. In Equation~\ref{eq:int_1}, the amplitude $|~\psi_{A}~|^2$ corresponds to $Pr(A)$ and $|~\psi_{B}~|^2$ corresponds to $Pr(B)$. So, what is the additional term, $ e^{i\theta_A}\psi_{A}~\cdot~e^{-i\theta_B}\psi_{B} + e^{i\theta_B}\psi_{B}~\cdot~e^{-i\theta_B}\psi_{A} $, that we derived in Equation~\ref{eq:int_1}? This term does not exist in classical probability theory and is called \emph{the interference term}.  The interference term can be rewritten like in Equation~\ref{eq:int_2}:

\begin{equation}
 e^{i\theta_A}\psi_{A}~\cdot~e^{-i\theta_B}\psi_{B} + e^{i\theta_B}\psi_{B}~ \cdot ~e^{-i\theta_A}\psi_{A} = |~\psi_{A}~| | \psi_{B} |e^{i ( \theta_A - \theta_B )} + |~\psi_{A}~| | \psi_{B} |e^{i ( \theta_B - \theta_A )}  
\label{eq:int_2}
\end{equation}

Knowing that, 
\[ cos( \theta_1 - \theta_2 ) = \frac{e^{i(\theta_1 - \theta_2)} + e^{i(-\theta_1 + \theta_2)}}{2} \] 
Then Equation~\ref{eq:int_2} can be rewritten as:

\begin{equation}
e^{i\theta_A} \psi_{A}~\cdot~e^{-i\theta_B}\psi_{B} + e^{i\theta_B}\psi_{B}~\cdot~e^{-i\theta_A}\psi_{A}  = 2 |~\psi_{A}~| | \psi_{B}|  \cos( \theta_A - \theta_B ) 
\label{eq:int_3}
\end{equation}

So, the complete quantum probability formula for the union of 2 mutually exclusive events is given by:

\begin{equation}
\psi_{AB = }|~e^{i\theta_A}\psi_{A} + e^{i\theta_B}\psi_{B} ~| ^2 =  |~\psi_{A}~|^2 + |~\psi_{B}~|^2  + 2~|~\psi_{A}~| ~| \psi_{B}|  \cos( \theta_A - \theta_B ) 
\label{eq:int_4}
\end{equation}
 
In the above formula, the angle $\theta_A - \theta_B$ corresponds to the phase of the inner product between $ |~\psi_{A}~|$ and $|~ \psi_{B} ~|$. 


Equation~\ref{eq:int_4} only computes the probability of the union of 2 mutually exclusive events. In classical probability theory, if we want to compute the probability of the union of $N$ mutually exclusive events, then we use a generalization of Equation~\ref{eq:union}, that is:

\begin{equation}
Pr( A_1 \cup A_2 \cup \dots \cup A_N ) = \sum_{i=1}^N Pr(A_i)
\label{eq:union_n}
\end{equation}

In quantum theory, we make an analogous calculation. In order to compute the probability of the union of $N$ mutually exclusive events, then one needs to generalize Equation~\ref{eq:qn_union}. The outcome is given by the following formula:

\[ Pr( A_1 \cup A_2 \cup \dots \cup A_N ) = | \psi_{A_1} + \psi_{A_2} + \dots + \psi_{A_N} |^2 = \left| \sum_{i=1}^N \psi_{A_i}  \right|^2\]

\begin{equation}
| e^{i\theta_1}\psi_{A_1} + e^{i\theta_2}\psi_{A_2} + \dots + e^{i\theta_{A_N}}\psi_N |^2 = \sum_{i=1}^N | \psi_{A_i} |^2 + 2 \sum_{i=1}^{N-1} \sum_{j=i+1}^{N} | \psi_{A_i} | | \psi_{A_j} | \cos( \theta_i - \theta_j )
\label{eq:union_qn_n}
\end{equation}

\subsection{Deriving the Interference Term from the Law of Total Probability}

This interference term can also be derived directly from the law of total probability. Suppose that events $A_1, A_2, \dots, A_N$ form a set of mutually disjoint events, such that their union is all in the sample space, $\Omega$, for any other event $B$. Then, the classical law of total probability can be formulated like in Equation~\ref{eq:law_total_prob_c}.

\begin{equation}
Pr(B) = \sum_{i=1}^{N} Pr(A_i) Pr(B | A_i)   \text{~~~~~~where:~~~~} \sum_{i = 1}^{N} Pr(A_i) = 1
\label{eq:law_total_prob_c}
\end{equation}

The quantum interference law of total probability can be derived through Equation~\ref{eq:law_total_prob_c} by applying Born's rule (Equation~\ref{eq:relation}). That is:

\begin{equation}
Pr(B) = \left| \sum_{x=1}^N e^{i\theta_x}\psi_{A_x} \psi_{B | A_x} \right|^2    \text{~~~~~~where:~~~~} \sum_{x= 1}^{N} \left| e^{i\theta_x} \psi_{A_x} \right|^2  = 1
\label{eq:total_law_q_1}
\end{equation}

For simplicity, we will expand Equation~\ref{eq:total_law_q_1} for $N=3$ and only later we will find the general formula for $N$ events:

\begin{equation}
Pr(B) = \left|  e^{i\theta_1} \psi_{A_1} \psi_{B | A_1} + e^{i\theta_2}\psi_{A_2} \psi_{B | A_2} +  e^{i\theta_3}\psi_{A_3} \psi_{B | A_3} \right|^2   
\label{eq:total_law_q_2}
\end{equation}

\begin{equation}
 Pr( B ) = \left(  e^{i\theta_1} \psi_{A_1} \psi_{B | A_1} + e^{i\theta_2}\psi_{A_2} \psi_{B | A_2} +  e^{i\theta_3}\psi_{A_3} \psi_{B | A_3}   \right) \left( e^{i\theta_1} \psi_{A_1} \psi_{B | A_1} + e^{i\theta_2}\psi_{A_2} \psi_{B | A_2} +  e^{i\theta_3}\psi_{A_3} \psi_{B | A_3}  \right)^{*}
\end{equation}

\begin{equation}
Pr(B) = \left(  e^{i\theta_1} \psi_{A_1} \psi_{B | A_1} + e^{i\theta_2}\psi_{A_2} \psi_{B | A_2} +  e^{i\theta_3}\psi_{A_3} \psi_{B | A_3} \right) \left(  e^{-i\theta_1} \psi_{A_1} \psi_{B | A_1} + e^{-i\theta_2}\psi_{A_2} \psi_{B | A_2} +  e^{-i\theta_3}\psi_{A_3} \psi_{B | A_3} \right)
\label{eq:total_law_q_3}
\end{equation}

\[Pr(B) = e^{i\theta_1} \psi_{A_1} \psi_{B | A_1}e^{-i\theta_1} \psi_{A_1} \psi_{B | A_1} + e^{i\theta_1} \psi_{A_1} \psi_{B | A_1} e^{-i\theta_2} \psi_{A_2} \psi_{B | A_2} + e^{i\theta_1} \psi_{A_1} \psi_{B | A_1} e^{-i\theta_3} \psi_{A_3} \psi_{B | A_3} + \]

\[ + e^{i\theta_2} \psi_{A_2} \psi_{B | A_2} e^{-i\theta_1} \psi_{A_1} \psi_{B | A_1}+e^{i\theta_2} \psi_{A_2} \psi_{B | A_2} e^{-i\theta_2} \psi_{A_2} \psi_{B | A_2} + e^{i\theta_2} \psi_{A_2} \psi_{B | A_2} e^{-i\theta_3} \psi_{A_3} \psi_{B | A_3} + \] 

\begin{equation}
+ e^{i\theta_3} \psi_{A_3} \psi_{B | A_3} e^{-i\theta_1} \psi_{A_1} \psi_{B | A_1}+e^{i\theta_3} \psi_{A_3} \psi_{B | A_3} e^{-i\theta_2} \psi_{A_2} \psi_{B | A_2} + e^{i\theta_3} \psi_{A_3} \psi_{B | A_3} e^{-i\theta_3} \psi_{A_3} \psi_{B | A_3}
\label{eq:total_law_q_4}
\end{equation}

Simplifying Equation~\ref{eq:total_law_q_4}, we obtain:

\begin{equation}
Pr(B) = \left| \psi_{A_1} \psi_{B | A_1} \right|^2 + \left| \psi_{A_2} \psi_{B | A_2} \right|^2 + \left| \psi_{A_3} \psi_{B | A_3} \right|^2 + Interference
\label{eq:total_law_q_5}
\end{equation}

In Equation~\ref{eq:total_law_q_5}, one can see that it is composed by the classical law of total probability and by an interference term. This interference term comes from Equation~\ref{eq:total_law_q_4} and corresponds to:

\[ Interference =  e^{i\theta_1 -i\theta_2 } \psi_{A_1} \psi_{B | A_1} \psi_{A_2} \psi_{B | A_2} + e^{i\theta_2} \psi_{A_2} \psi_{B | A_2} e^{-i\theta_1} \psi_{A_1} \psi_{B | A_1} + \]

\[ +  e^{i\theta_1-i\theta_3} \psi_{A_1} \psi_{B | A_1} \psi_{A_3} \psi_{B | A_3}  + e^{i\theta_3 -i\theta_1} \psi_{A_3} \psi_{B | A_3} \psi_{A_1} \psi_{B | A_1} +  \] 

\begin{equation}
 +  e^{i\theta_2 -i\theta_3} \psi_{A_2} \psi_{B | A_2} \psi_{A_3} \psi_{B | A_3} + e^{i\theta_3 -i\theta_2} \psi_{A_3} \psi_{B | A_3} \psi_{A_2} \psi_{B | A_2}  
\label{eq:total_law_q_6}
\end{equation}

Knowing that
\[ \cos( \theta ) = \frac{ e^{i\theta} + e^{-i\theta} }{2}  \Rightarrow  \cos( \theta_1 - \theta_2) = \frac{e^{i\theta_1- i\theta_2} + e^{i\theta_2- i\theta_1}}{2} \]
Then, Equation~\ref{eq:total_law_q_5} becomes

\begin{equation}
\begin{split}
Pr(B) = \left| \psi_{A_1} \psi_{B | A_1} \right|^2 + \left| \psi_{A_2} \psi_{B | A_2} \right|^2 + \left| \psi_{A_3} \psi_{B | A_3} \right|^2 
+ 2\psi_{A_1} \psi_{B | A_1} \psi_{A_2} \psi_{B | A_2} \cos(\theta_1 -\theta_2) +\\ 
+2\psi_{A_1} \psi_{B | A_1} \psi_{A_3} \psi_{B | A_3}\cos(\theta_1 -\theta_3) + 2\psi_{A_2} \psi_{B | A_2} \psi_{A_3} \psi_{B | A_3} \cos(\theta_2 - \theta_3)
\end{split}
\label{eq:total_law_q_7}
\end{equation}

Generalizing Equation~\ref{eq:total_law_q_7} for $N$ events, the final probabilistic interference formula, derived from the law of total probability, is given by:

\begin{equation}
Pr(B) = \sum_{i = 1}^N  \left| \psi_{A_i} \psi_{B | A_i} \right|^2 + 2\sum_{i=1}^{N-1}\sum_{j= i + 1}^N \psi_{A_i} \psi_{B | A_i} \psi_{A_j} \psi_{B | A_j} \cos(\theta_i -\theta_j)
\label{eq:total_law_q_final}
\end{equation}

Following Equation~\ref{eq:total_law_q_final}, when $ \cos( \theta_i - \theta_j) $  equals zero, then it is straightforward that quantum probability theory converges to its classical counterpart, because the interference term will be zero. 

For non-zero values, Equation~\ref{eq:total_law_q_final} will produce interference effects that can affect destructively the classical probability ( when interference term in smaller than zero ) or  constructively ( when it is bigger than zero ). Additionally,  Equation~\ref{eq:total_law_q_final} will lead to a large amount of $\theta$ parameters when the number of events increases. For $N$ binary random variables, we will end up with $2^{N}$ parameters to tune.

\section{The Role of the Interference Term in the Literature}\label{sec:theta_lit}

In the 20th century, the physicist Max Born proposed a problem related to quantum probabilities, which was later known as the \emph{The Inverse Born Problem}. The problem consisted in constructing a probabilistic representation of data from different sources (physics, psychology, economy, etc), by a complex probability amplitude, which could match Born's rule (already presented in Equation~\ref{eq:relation}). 

The probabilistic interference formula for the law of total probability, derived in the previous section (Equation~\ref{eq:total_law_q_final}), can be seen as an answer to the Inverse Born Problem. The most important works in the literature that contributed for the derivation of this interference term, through the law of total probability, correspond to the works of A. Khrennikov~\cite{Khrennikov09quantumlike,Khrennikov06,Khrennikov09sure,Haven13,Khrennikov05Classical,Conte09}. These authors address the interference term as $\left| \lambda(B|A)\right|$. In the situations where $ \left| \lambda(B|A) \right| \le 1 $, then one can apply the trigonometric formula derived in Equation~\ref{eq:total_law_q_final}. However, there are some data where this condition is not verified. Therefore, when $ \left| \lambda(B|A) \right| \ge 1 $, the authors propose the usage of a hyperbolic interference term, to act like an upper boundary in order to constraint the probability value to a maximum value of $1$. This would require the usage of Hyperbolic Hilbert Spaces instead of the complex ones.

In this paper, we argue that there is no need to represent probabilities in a Hyperbolic Hilbert Space in order to avoid non-probability values. Since we will be dealing with probabilistic graphical models, we will always be required to normalise the probability amplitudes when performing probabilistic inferences. For this work, we follow the same probabilistic paradigm used in traditional Bayesian Networks. Thus, we constrain Equation~\ref{eq:total_law_q_final_norm} and Equation~\ref{eq:union_qn_norm} by a normalisation factor $\alpha$ that will guarantee that the computed values will always be probabilities lesser or equal than one. This normalisation factor corresponds to Feynman's conjecture~\cite{Feynman65} that an electron can follow any path. Thus, in order to compute the probability $Pr(B)$ that a particle ends up at a point $B$ , one must sum over all possible paths that the particle can go through. Since the interference term can lead to estimations with values higher than $1$, then it is necessary to normalise in order to obtain a probability value.

\begin{equation}
Pr(B) = \alpha \left[ \sum_{i = 1}^N  \left| \psi_{A_i} \psi_{B | A_i} \right|^2 + 2\sum_{i=1}^{N-1}\sum_{j=i+1}^N \psi_{A_i} \psi_{B | A_i} \psi_{A_j} \psi_{B | A_j} \cos(\theta_i -\theta_j) \right] \text{~~~~where~~~~~} \alpha = \frac{1}{Pr(B) + Pr( \neg B)}
\label{eq:total_law_q_final_norm}
\end{equation}

\begin{equation}
 Pr( A_1 + \dots + A_N ) =  \alpha \left[  \sum_{i=1}^N | \psi_{A_i} |^2 + 2 \sum_{i=1}^{N-1} \sum_{j=i+1}^{N} | \psi_{A_i} | | \psi_{A_j} | \cos( \theta_i - \theta_j ) \right]
\label{eq:union_qn_norm}
\end{equation}

\[  \text{~where~} \alpha = \frac{1}{Pr(A_1 + \dots + A_N ) + Pr( \neg A_1 + \dots + \neg A_N )}\]

Through the triangular inequality, we can also prove that the proposed interference term is also always positive. This way, the axioms of probability theory that state that $ 0 \le Pr(A) \le 1 $ will always be satisfied. By applying the triangular inequality, one can easily demonstrate that Equation~\ref{eq:union_qn_norm} and, consequently, Equation~\ref{eq:total_law_q_final_norm} have always to be positive.

Through the triangular inequality, Equation~\ref{eq:union_qn_norm} can be related to:

\begin{equation}
| \psi_{A_1} + \psi_{A_2} + \dots + \psi_{A_N} |^2 \le ( | \psi_{A_1}| + |\psi_{A_2}| + \dots + |\psi_{A_N} |)^2
\end{equation}

\begin{equation}
\sum_{i=1}^N | \psi_{A_i} |^2 + 2 \sum_{i=1}^{N-1} \sum_{j=i+1}^{N} | \psi_{A_i} | | \psi_{A_j} | \cos( \theta_i - \theta_j ) \le \sum_{i=1}^N | \psi_{A_i} |^2 + 2 \sum_{i=1}^{N-1} \sum_{j=i+1}^{N} | \psi_{A_i} | | \psi_{A_j} | 
\end{equation}

\begin{equation}
 \sum_{i=1}^{N-1} \sum_{j=i+1}^{N} | \psi_{A_i} | | \psi_{A_j} | \cos( \theta_i - \theta_j ) \le \sum_{i=1}^{N-1} \sum_{j=i+1}^{N} | \psi_{A_i} | | \psi_{A_j} | 
\end{equation}

\begin{equation}
 - \sum_{i=1}^{N-1} \sum_{j=i+1}^{N} | \psi_{A_i} | | \psi_{A_j} | \cos( \theta_i - \theta_j ) + \sum_{i=1}^{N-1} \sum_{j=i+1}^{N} | \psi_{A_i} | | \psi_{A_j} | \ge 0 
\end{equation}

\begin{equation}
 \sum_{i=1}^{N-1} \sum_{j=i+1}^{N} | \psi_{A_i} | | \psi_{A_j} | \left[  1 - \cos(\theta_i - \theta_j)  \right] \ge 0
\end{equation}

The maximum value that $\cos(\theta_i - \theta_j)$ can have is $1$ and the minimum value is $-1$. So, the minimum value that the term $1 - \cos(\theta_i - \theta_j) $ can have is $0$ (when the cosine achieves its maximum value). Given that $\psi_{A_1},~\psi_{A_2},~\dots, ~\psi_{A_N}$ are always positive numbers, then it is straightforward that Equation~\ref{eq:total_law_q_final_norm} has to be always positive.

\section{Classical vs Quantum Bayeisan Network to Model Two Stage Gambles}\label{sec:classic_vs_quantum_gamble}

\begin{figure}[ht]
	\parbox{.48\linewidth}{
	\centering
	\includegraphics[scale=0.45]{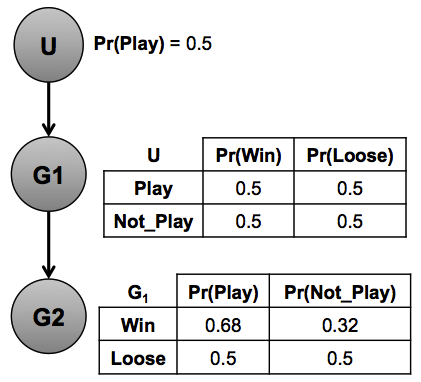}
	\caption{Classical Bayesian Network for the Two-Stage Gambling Game}
	\label{fig:classical}
	}
	\hfill
	\parbox{.48\linewidth}{
	\centering
	\includegraphics[scale=0.45]{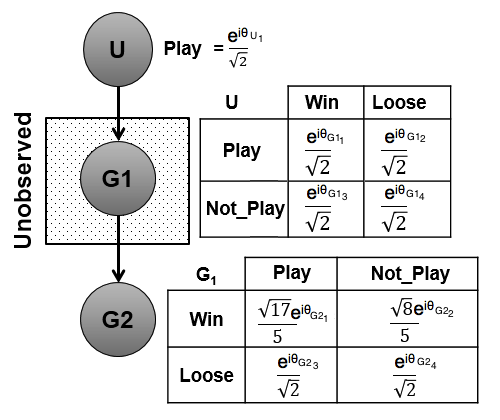}
	\caption{Quantum Bayesian Network for the Two-Stage Gambling Game}
	\label{fig:quantum}	
	}
\end{figure}

\subsection{Classical Bayesian Networks}

A classical Bayesian Network can be defined by a directed acyclic graph structure in which each node represents a different random variable from a specific domain and each edge represents a direct influence from the source node to the target node. The graph represents independence relationships between variables and each node is associated with a conditional probability table which specifies a distribution over the values of a node given each possible joint assignment of values of its parents.  This idea of a node depending directly from its parent nodes is the core of Bayesian Networks. Once the values of the parents are known, no information relating directly or indirectly to its parents or other ancestors can influence the beliefs about it~\cite{koller09prob}.

\subsubsection{Classical Conditional Independece}

Associated to Bayesian Networks there is always the concept of conditional independence. Two random variables $X$ and $Y$ are conditionally independent given a third random variable $Z$ if and only if they are independent in their conditional probability distribution given $Z$. In other words, $X$ and $Y$ are conditionally independent given $Z$, $(X=x \perp Y=y | Z)$, if and only if, given any value of $Z$, the probability distribution of $X$ is the same for all values of $Y$ and the probability distribution of $Y$ is the same for all values of $X$.

This means that an independence statement over random variables is a universal quantification over all possible values of random variables~\cite{koller09prob}. Therefore, a probability distribution $Pr$ satisfies $(X \perp Y | Z)$ if and only if:
\begin{equation}
Pr(X,Y | Z) = Pr(X | Z) Pr(Y | Z) 
\label{eq:classical_condi}
\end{equation}

\subsubsection{Classical Random Variables}

In classical probability theory, a random variable $X$ is defined by a function that associates a value to each outcome in the sample space $\Omega$, $X: \Omega \rightarrow \mathbb{R}$. 

\subsubsection{Example of Application in the Two-Stage Gambling Game}

In the two-stage gambling game, the random variables correspond to the nodes and their respective conditional probability tables of the Bayesian Network in Figure~\ref{fig:classical}. That is, the variable $U$ corresponds to a player willing or not to participate in the game. Variable $G_1$ corresponds to the variable winning or losing the first gamble. Variable $G_2$ corresponds to a player playing or not the second gamble. 

\subsubsection{Classical Full Joint Distributions} 

In classical probability theory, the full joint distribution over a set of $n$ random variables $\chi$ $=$ $\{$ $X_1,$ $X_2,$ $...,$ $X_n\}$ defined over the same sample space, $Pr(X_1, X_2, ..., X_n)$, is the distribution that assigns probabilities to events that are specified in terms of these random variable~\cite{koller09prob}.Then, the full joint distribution of a Bayesian Network, where $X$ is the list of variables, is given by~\cite{russel10}: 

\begin{equation}
Pr( X_1, \dots, X_n ) = \prod_{i=1}^n  Pr( X_i | Parents(X_i) ) 
\label{eq:joint}
\end{equation}

\subsubsection{Example of Application in the Two-Stage Gambling Game}

Using Equation~\ref{eq:joint}, the full joint distribution of Figure~\ref{fig:classical} corresponds to the calculations discriminated in Tables~\ref{tab:full_joint}.

\begin{table}[h!]
\centering
\begin{tabular}{l l l  | c}
{\bf U}		& {\bf G$_1$}		& {\bf G$_2$}	& {\bf Pr(U, G$_1$, G$_2$) }   \\
\hline
Play			& Win	 		& Play				& 0.5 $\times$ 0.5 $\times$ 0.68 = 0.17	\\
Play			& Win			& Not\_Play			& 0.5 $\times$ 0.5 $\times$ 0.32 = 0.08	\\
Play			& Lose		     	& Play				& 0.5 $\times$ 0.5 $\times$ 0.5 = 0.125	\\
Play			& Lose		     	& Not\_Play			& 0.5 $\times$ 0.5 $\times$ 0.5 = 0.125	\\
Not\_Play		& Win		     	& Play				& 0.5 $\times$ 0.5 $\times$ 0.68 = 0.17	\\
Not\_Play		& Win		     	& Play				& 0.5 $\times$ 0.5 $\times$ 0.32 = 0.08	\\
Not\_Play		& Lose		     	& Play				& 0.5 $\times$ 0.5 $\times$ 0.5 = 0.125	\\
Not\_Play		& Lose		     	& Play				& 0.5 $\times$ 0.5 $\times$ 0.5 = 0.125	\\
\end{tabular}
\caption{Fulll joint distribution of the Bayesian Newtwork in Figure~\ref{fig:classical}}
\label{tab:full_joint}
\end{table}

\subsubsection{Classical Marginalization}

Given a query random variable $X$ and let $Y$ be the unobserved variables in the network, the marginal distribution of $X$ is simply the probability distribution of $X$ averaging over the information about $Y$. The marginal probability for discrete random variables, can be defined by Equation~\ref{eq:marginal_prob}. The summation is over all possible $y$, i.e., all possible combinations of values of the unobserved variables $y$. The term $\alpha$ corresponds to a normalization factor for the distribution $Pr( X )$~\cite{russel10}.

\begin{equation}
\begin{split}
Pr(X=x) = \alpha \sum_y Pr(X = x, Y = y) = \alpha \sum_y Pr(X = x | Y = y) Pr(Y = y)\text{, where } \alpha = \frac{1}{ \sum_{x \in X} Pr(X = x) } 
\end{split}
\label{eq:marginal_prob}
\end{equation}

\subsubsection{Example of Application in the Two-Stage Gambling Game}

After computing the full joint distribution, we need to sum out all the variables that are unknown, in this case, the variable corresponding to the outcome of the first gamble: $G_1$. This is achieved by applying the marginal probability formula in Equation~\ref{eq:marginal_prob}.

\[ Pr( G_2 = Play ~|~ U = Play ) = \alpha \sum_{g \in G_1} Pr( U = Play, G_1 = g, G_2 = Play )  \] 

\[ Pr(G_2 = Play ~|~ U = Play) = \alpha ~ Pr(U = Play) \sum_{ g \in G_1} Pr(G_1 = g ~ | ~ U = Play) ~ Pr( G_2 = Play ~|~ G_1 = g) \]
\[ Pr(G_2 = Play ~|~ U = Play) = \alpha ~ Pr(U = Play) ~ [ Pr(G_1 = win ~ | ~ U = Play) ~ Pr( G_2 = Play ~|~ G_1 = win) ~+~ \]
\[    										~~~~~~~~~~~~~~+~Pr(G_1 = lose ~ | ~ U = Play) ~ Pr( G_2 = Play ~|~ G_1 = lose) ]  \]

\[ Pr(G_2 = Play ~|~ U = Play) = \alpha 0.295 \]
\[ Pr(G_2 = Not\_Play ~|~ U = Play) = \alpha 0.205 \] 

The parameter $\alpha$ corresponds to a normalisation factor and, for this example, is given by:
\[ \alpha =  \frac{1}{Pr(G_2 = Play ~|~ U = Play) + Pr(G_2 = Not\_Play ~|~ U = Play)} = \frac{1}{0.295 + 0.205} = \frac{1}{0.5} \]

So, the final normalised classical probabilities for the two-stage gambling game correspond to:
\[ Pr(G_2 = play ~|~ U = play) =  0.59 \]
\[ Pr(G_2 = Not\_Play ~|~ U = Play) = 0.41 \]

The probabilities computed are not in accordance with the probabilistic findings reported by~\cite{Tversky92}, because it was empirically observed that $Pr(G_2 = play ~|~ U = play) =  0.42$. Therefore, a classical Bayesian Network can never be used to model such experiments, because of the relation already presented in Equation~\ref{eq:relation_observed}.

\subsection{The Quantum Interference Bayesian Network}

Following the work of~\cite{Leifer08}, a Quantum Bayesian Network can be defined by a pair (G,$\rho_v$), where G = (V,E) is a directed acyclic graph, and each vertex $v \in V$ is associated with a quantum system with a Hilbert space $H_v$ and $\rho_v$ is a quantum state on $H_V = H_{v1} \otimes H_{v2} \otimes \dots \otimes H_{vn}$. The state $\rho_v$ satisfies the same conditional independence constraints as in a classical Bayesian Network. Note that the definition of a classical Bayesian Network can be directly obtained by replacing the word quantum system by random variable.

The symbol $\otimes$ is defined by \emph{tensor product} and corresponds to a mathematical method that enables the construction of a Hilbert space from the combination of individual Hilbert spaces. Suppose that we have 2 different 2-dimensional Hilbert spaces $H_x$ and $H_y$, where $H_x = \{ x_1, x_2 \}$ and $H_y = \{ y_1, y_2 \}$.  Then, their tensor product would be:

\[ \left[ \begin{matrix} x_1 \\ x_2 \end{matrix} \right]  \otimes \left[ \begin{matrix} y_1 \\ y_2 \end{matrix} \right] = \left[ \begin{matrix} x_1 ~y_1\\ x_1 ~y_2 \\ x_2 ~y_1\\ x_2 ~y_2 \end{matrix} \right] \] 

\subsubsection{Quantum Random Variables}

 In quantum theory, random variables are associated to a set of $N$ quantum systems $V = \{ v_1, v_2, ..., v_N \}$, each associated with a Hilbert space with a specific dimension\cite{Leifer08}. Consequently, all values contained in the conditional probability tables associated to the random variables are complex numbers.

\subsubsection{Example of Application in the Two-Stage Gambling Game}

Each node  on the Bayesian Network in Figure~\ref{fig:quantum} can be seen as a subsystem belonging to a specific Hilbert space. For instance, the first node $U$ can be represented in a Hilbert subspace $H_u$ in the following way:
\[ U = e^{i\theta_U}\frac{1}{\sqrt{2}}Play + e^{i\theta_U}\frac{1}{\sqrt{2}}Not\_Play \]
Where $Play$ and $Not\_Play$ are column vectors corresponding to the basis of the subspace $H_u$:

\[ Play = \left[ \begin{matrix} 1 \\ 0 \end{matrix} \right] ~~~~ Not\_Play = \left[ \begin{matrix} 0 \\ 1 \end{matrix} \right]\]

Since our goal is to compare the quantum Bayesian Network with its classical counterpart, we will convert the conditional probability tables in Figure~\ref{fig:classical} into conditional amplitude tables. That is, we simply convert classical probabilities into complex amplitudes through the relation in Equation~\ref{eq:relation}. 
\begin{equation}
Pr(A) = | e^{i\theta_A}\psi_A |^2 \rightarrow \psi_A = e^{i\theta_A}\sqrt{Pr(A)} 
\end{equation}

\subsubsection{Quantum State}

The representation of a general state in a Bayesian Network can be described by a bipartite state. Suppose
that $H = H_X \otimes H_Y \otimes H_Z$ is a Hilbert space defined by the composition of three Hilbert spaces
$H_X$, $H_Y$ and $H_Z$. Then, a quantum state $S_{XYZ}$ is designated bipartite if it can be specified with respect to the random variables $X$, $Y$ and $Z$. For $X_i$, $Y_j$ and $Z_k$ as basis in $H_X$, $H_Y$ and $H_Z$, respectively, the bipartite state is given by Equation \ref{eq:bipartite_state}.

\begin{equation}
S_{XYZ} = \sum_{i,j,k}  \psi_{Xi} \psi_{Yj} \psi_{Zk} X_i \otimes Y_j \otimes Z_k
\label{eq:bipartite_state}
\end{equation}

In Equation~\ref{eq:bipartite_state}, $\psi_{Xi} \psi_{Yj} \psi_{Zk}$ corresponds to the amplitudes of states $X_i$, $Y_j$ and $Z_k$, respectively.  The states $X_i$, $Y_j $ and  $Z_k $ correspond to column vectors representing basis vectors:

\[ X_0 = Y_0 = Z_0 =  \left[  \begin{matrix}1\\ 0    \end{matrix} \right]  \text{,~~~~} X_1 = Y_1 = Z_1 = \left[  \begin{matrix}0\\ 1 \\  \end{matrix} \right]  \]

\subsubsection{Example of Application in the Two-Stage Gambling Game}

The general quantum state represented by the Bayesian Network in Figure~\ref{fig:quantum} is given by quantum bipartite states. Reformulating Equation~\ref{eq:bipartite_state} for the problem of two step gambles, we obtain:

\begin{equation}
 S_{U,G1, G2} = \sum_{ijk}  \psi_{Ui} \psi_{G1j} \psi_{G2k}U_i \otimes G1_j \otimes G2_k
\label{eq:gamble_bipartite}
\end{equation}

For simplicity, we will write $  \psi_{Ui} \psi_{G1j} \psi_{G2k}$ as $\psi_{ijk}$. So, expanding Equation~\ref{eq:gamble_bipartite},

\[  S_{U,G1, G2} =  \psi_{000}~U_0G1_0G2_0 e^{i (\theta_{U0} + \theta_{G10} + \theta_{G20}) }  +  \psi_{001}~U_0 G1_0 G2_1e^{i (\theta_{U0} + \theta_{G10} + \theta_{G21}) } +  \psi_{010}~U_0 G1_1 G2_0e^{i (\theta_{U0} + \theta_{G11} + \theta_{G20}) }  + \]   
\[ + \psi_{011}~U_0 G1_1 G2_1e^{i (\theta_{U0} + \theta_{G11} + \theta_{G21}) } + \psi_{100}~U_1 G1_0 G2_0e^{i (\theta_{U1} + \theta_{G10} + \theta_{G20}) }  +  \psi_{101}~U_1 G1_0 G2_1e^{i (\theta_{U1} + \theta_{G10} + \theta_{G21}) } +\]
\begin{equation}
+  \psi_{110}~U_1 G1_1 G2_0e^{i (\theta_{U1} + \theta_{G11} + \theta_{G20}) } +  \psi_{111}~U_1 G1_1 G2_1e^{i (\theta_{U1} + \theta_{G11} + \theta_{G21}) } 
\label{eq:wtv} 
\end{equation} 

Note that $U_iG1_jG2_k$ are basis column vectors representing the axis of the Hilbert Space, 
\[ U_0G1_0G2_0 = \left[ \begin{matrix} ~ 1 ~ \\ ~ 0 ~ \\ ~ 0 ~ \\ ~ 0 ~ \\ ~ 0 ~ \\ ~ 0 ~ \\ ~ 0 ~ \\~ 0 ~ \end{matrix} \right], ~ U_0G1_0G2_1 = \left[ \begin{matrix} ~ 0 ~ \\ ~ 1 ~ \\ ~ 0 ~ \\ ~ 0 ~ \\ ~ 0 ~ \\ ~ 0 ~ \\ ~ 0 ~ \\~ 0 ~ \end{matrix} \right],~U_0G1_1G2_0 = \left[ \begin{matrix} ~ 0 ~ \\ ~ 0 ~ \\ ~ 1 ~ \\ ~ 0 ~ \\ ~ 0 ~ \\ ~ 0 ~ \\ ~ 0 ~ \\~ 0 ~ \end{matrix} \right], \dots ~U_1G1_1G2_1 = \left[ \begin{matrix} ~ 0 ~ \\ ~ 0 ~ \\ ~ 0 ~ \\ ~ 0 ~ \\ ~ 0 ~ \\ ~ 0 ~ \\ ~ 0 ~ \\~ 1 ~ \end{matrix} \right],\]

Following the quantum Bayesian Network in Figure~\ref{fig:quantum}, we compute the values of $\psi_{Ui} \psi_{G1j} \psi_{G2k}$ by multiplying the correspondent values in the conditional probability tables. For example, $\psi_{000}$ corresponds to the product of the variables $U=Play$ with $G1=Win~ |~U = Play$ with $G2 = Play~|~ G1 = Win$. On the other hand, the entry  $\psi_{110}$ corresponds to the product of the variables $U=Not\_Play$ with $G1=Lose~ |~U = Not \_ Play$ with $G2 = Play~|~ G1 = Lose$. Replacing Equation~\ref{eq:wtv} by the conditional probability values in Figure~\ref{fig:quantum}, we obtain:

\[  S_{U,G1, G2} = e^{i (\theta_{U0} + \theta_{G10} + \theta_{G20}) } \frac{1}{\sqrt{2}}\frac{1}{\sqrt{2}}\frac{\sqrt{17}}{5}U_0G1_0G2_0 + e^{i (\theta_{U0} + \theta_{G10} + \theta_{G21}) }\frac{1}{\sqrt{2}}\frac{1}{\sqrt{2}}\frac{\sqrt{8}}{5}U_0 G1_0 G2_1 + e^{i (\theta_{U0} + \theta_{G11} + \theta_{G20}) }  \frac{1}{\sqrt{2}}\frac{1}{\sqrt{2}}\frac{1}{\sqrt{2}}U_0 G1_1 G2_0 +  \]    
\[ + e^{i (\theta_{U0} + \theta_{G11} + \theta_{G21}) }\frac{1}{\sqrt{2}}\frac{1}{\sqrt{2}}\frac{1}{\sqrt{2}}U_0 G1_1 G2_1 + e^{i (\theta_{U1} + \theta_{G10} + \theta_{G20}) }\frac{1}{\sqrt{2}}\frac{1}{\sqrt{2}}\frac{\sqrt{17}}{5} U_1 G1_0 G2_0 +  e^{i (\theta_{U1} + \theta_{G10} + \theta_{G21}) } \frac{1}{\sqrt{2}}\frac{1}{\sqrt{2}}\frac{\sqrt{8}}{5}U_1 G1_0 G2_1  \]

\[  +e^{i (\theta_{U1} + \theta_{G11} + \theta_{G20}) } \frac{1}{\sqrt{2}}\frac{1}{\sqrt{2}}\frac{1}{\sqrt{2}}U_1 G1_1 G2_0 + e^{i (\theta_{U1} + \theta_{G11} + \theta_{G21}) }\frac{1}{\sqrt{2}}\frac{1}{\sqrt{2}}\frac{1}{\sqrt{2}}U_1 G1_1 G2_1\]

\[  S_{U,G1, G2} = 0.4123 ~ U_0G1_0G2_0 + 0.2828 ~ U_0 G1_0 G2_1 + 0.3536 ~U_0 G1_1 G2_0 + 0.3536 ~ U_0 G1_1 G2_1 + \]    
\[+ 0.4123 ~U_1 G1_0 G2_0 + 0.2828\frac{\sqrt{8}}{5}U_1 ~ G1_0 G2_1 + 0.3536 ~ U_1 G1_1 G2_0 + 0.3536 ~U_1 G1_1 G2_1 \]

Note that the sum of the squares of all probability amplitudes $\psi_{Ui} \psi_{G1j}\psi_{G2k}$ sum to 1,
\[ \sum_{ijk} \left| \psi_{Ui} \psi_{G1j} \psi_{G2k} \right|^2 = 1\]

This bipartite state represents a quantum superposition over all possible states. We can think of this as various wave functions that are occurring at the same time. 

\subsubsection{Quantum Full Joint Distribution}

In quantum probability theory, a full joint distribution is given by a \emph{density matrix}. This matrix provides the probability distribution of all states that a Bayesian Network can have. In our quantum Bayesian Network model, the density matrix $\rho$ corresponds to the multiplication of the bipartite state described in Equation~\ref{eq:bipartite_state} with itself (the symbol $\dagger$ corresponds to the conjugate transpose):

\begin{equation}
\rho = S_{XYZ}^{~} S_{XYZ}^{\dagger}
\label{eq:density}
\end{equation}

The reason why one multiplies the same state with itself is to obtain the probability value out of the amplitude. Note that the bipartite state contains amplitudes instead of probability values. Knowing that the probability value is obtained by taking the squared magnitude of the amplitude, then, by multiplying a state with its conjugate transpose,  one can obtain the full joint probability distribution.

\subsubsection{Example of Application in the Two-Stage Gambling Game}

From the bipartite state, one can compute the density matrix by applying Equation~\ref{eq:density} to this calculation. The density matrix will be useful for later calculations and enables the calculation of the probability distribution of the bipartite state.

\begin{equation}
\rho_{UG1G2} = S_{U G1 G2}  ~\cdot ~ S_{U G1 G2 }^{\dagger}
\label{eq:density_game}
\end{equation}

In this two step gambling game, Equation~\ref{eq:density_game} produces an 8 x 8 density matrix:

\begin{equation}
\footnotesize
\rho_{UG1G2} =  \left[ ~\begin{matrix} |\psi_{000}|^2	&0 			&0 			&\dots 			&0 \\
			   0 			& |\psi_{001}|^2	&0			&\dots			&0\\	
			   0			&0			&|\psi_{010}|^2	&\dots			&0\\
			  \vdots		&\vdots		& \vdots		&\ddots			&\vdots \\
			  0			&0			&0			&\dots			& |\psi_{111}|^2
   \end{matrix} ~ \right]  ~=~ \left[ ~\begin{matrix} 
0.1700	&0		&0		&0		&0		&0		&0		&0 \\
0		&0.0800	&0		&0		&0		&0		&0		&0 \\
0		&0		&0.1250	&0		&0		&0		&0		&0 \\
0		&0		&0		&0.1250	&0		&0		&0		&0 \\
0		&0		&0		&0		&0.1700	&0		&0		&0 \\
0		&0		&0		&0		&0		&0.0800	&0		&0 \\
0		&0		&0		&0		&0		&0		&0.1250	&0 \\
0		&0		&0		&0		&0		&0		&0		&0.1250 \\
			 
   \end{matrix} ~ \right]  
\end{equation}

\subsubsection{Quantum Marginalization}

The concept of quantum marginalization is analogous to the one in classical probability theory. Given two quantum random variables $X$ and $Y$, the general idea is to compute the average of the probability distribution of $X$ over the information about $Y$. This is performed by using the \emph{partial trace} operator, which basically consists in accessing certain positions of the density matrix $\rho$. 

\begin{equation}
X_{ij} = \alpha  \sum_{y \in Y}  \rho[iy, jy]  
\label{eq:quantum_marginalization}
\end{equation}

In Equation~\ref{eq:quantum_marginalization}, the parameter $\alpha$ corresponds to the normalization factor, which is also present in the classical Bayesian Network inference.

\subsubsection{Example of Application in the Two-Stage Gambling Game}

Considering the two-stage gambling game that we are analyzing, imagining that we want to determine the probability of a participant playing the second gamble, $Pr(G2_{00})$, given that we verified that the player participated in the game in the first place ($u=0$). That is, we will be summing out variable $G1$. Through Equation~\ref{eq:quantum_marginalization}, one would proceed in the following way:

\[ G2_{ij} = G2_{00} = \alpha \sum_{g \in G1}  \rho[ugi, ugj] = \alpha ( \rho[000, 000] + \rho[010, 010] )= \alpha 0.2950 \rightarrow 0.59 \] 

And in the same way for $G2_{11}$:
 \[ G2_{11} = \alpha \sum_{g \in G1, u \in U}  \rho[ugi, ugj] = \alpha ( \rho[001, 001] + \rho[011, 011]  )= \alpha 0.2050 \rightarrow 0.41 \] 

Note that the indexes of the density matrix are encoded. The assignment $000$ corresponds to index $0$, the assignments $001$ to index $1$, $\dots$, and the assignment $111$ corresponds to the index $8$.

The normalised results correspond to the same probabilities obtained in classical theory. Therefore, we need to incorporate the interference terms found in cognitive psychology into the quantum marginalization formula in order to obtain different results.

\subsubsection{Quantum Marginalization with Interference}

A quantum interference effect will occur when performing the marginalization of a random variable. If we \emph{do not observe} a random variable, then it can be in represented by superposition and perform interferences in other random variables, changing the final outcome.

The quantum interference marginalization formula proposed in this work consists in merging the equation that presents the interference term (Equation~\ref{eq:union_qn_n}) with the formula of quantum marginalization (Equation~\ref{eq:quantum_marginalization}). This leads to Equation~\ref{eq:form}:

\[ X_{ij} = \alpha \left|  \sum_{y \in Y}  \rho[iy, jy]   \right| ^2 \]

\begin{equation}
X_{ij} = \alpha \left( \sum_i^N | \psi_i |^2  + 2 \sum_i^{N-1}\sum_{j = i + 1}^{N} | \psi_i| |\psi_j| cos( \theta_i - \theta_j  ) \right)\text{,~~~where } | \psi_i |^2 = \sum_{y \in Y} \rho[iy, jy]  
\label{eq:form}
\end{equation}

\subsubsection{Example of Application in the Two-Stage Gambling Game}

Following the experiment performed by~\cite{Tversky92}, we want to determine the probability of a participant playing the second gamble, given that (s)he does not know the outcome of the first gamble.

Through Equation~\ref{eq:form}, this can be computed in the following way:

\[  G2_{ij} = \alpha \left|  \sum_{u \in U, g \in G1}  \rho[iug, jug]  \cdot \right| ^2 \]

Given that we want to know the probability of a participant playing the second gamble, then we know for certain that (s)he accepted to play. Thus, we will fix the variable $U$, representing that a participant accepted to play the game in the first place, $U=0$:

\begin{equation}
G2_{ij} = \alpha \left|  \sum_{g \in G1}  \rho[i0g, j0g]   \right| ^2
\label{eq:partial_trace_game}
\end{equation}

Expanding Equation~\ref{eq:partial_trace_game},
\[ G2_{00} =   \alpha \left| \rho[000, 000]     + \rho[010, 010]    \right| ^2 \]

\begin{equation}
G2_{00}  = \alpha \left|  0.17 + 0.125  \right|^2 = \alpha (0.17 + 0.125+ 2\sqrt{0.17}\sqrt{0.125}\cos(\theta_1 - \theta_2) )
\label{eq:G00}
\end{equation}

Computing the probability of a participant deciding to not play the second gamble, given that he does not know the outcome of the first play, we obtain:

\[ G2_{11} =    \alpha \left| \rho[001, 001]     + \rho[011, 011]    \right| ^2 \]

\begin{equation}
 G2_{11} = \alpha \left|  0.08 + 0.125  \right|^2 = \alpha (0.08 + 0.125 + 2\sqrt{0.08}\sqrt{0.125}\cos(\theta_1 - \theta_2) ) 
\label{eq:G11}
\end{equation}
The normalization factor $\alpha$ is computed by summing the results when $G2 = Play$ and $G2 = Not\_Play$, that is, summing Equation~\ref{eq:G00} with Equation~\ref{eq:G11}:

\[ \alpha = \frac{1}{0.5 + 2\sqrt{0.17}\sqrt{0.125}\cos(\theta_1 - \theta_2) + 2\sqrt{0.08}\sqrt{0.125}\cos(\theta_1 - \theta_2)  } = \frac{1}{0.5 + 0.4915\cos(\theta_1 - \theta_2)} \]

Then, the normalized results are given by:

\begin{equation}
G2_{00}  = \frac{ 0.2950 + 2\sqrt{0.17}\sqrt{0.125}\cos(\theta_1 - \theta_2) }{ 0.5 + 0.4915\cos(\theta_1 - \theta_2) } 
\label{eq:g00_norm}
\end{equation}

\begin{equation}
G2_{11} = \frac{ 0.2050 + 2\sqrt{0.08}\sqrt{0.125}\cos(\theta_1 - \theta_2) } { 0.5 + 0.4915\cos(\theta_1 - \theta_2) }
\end{equation}

The aim of this quantum interference Bayesian Network is to simulate the averaged results reported in Table~\ref{tab:game_results}. More specifically, we are interested in simulating the value corresponding to the probability of the participant playing the second gamble, given that (s)he does not know the outcome of the first one. In Table~\ref{tab:game_results}, this value corresponds to $0.42$ and cannot be obtained through classical probability theory, since the law of total probability is violated. In our model, we can tune the angle $\theta$ of the interference term in order to obtain such results. Calculations showed that in order to achieve a probability of $42\%$, $\cos(\theta)$ must be equal to $-0.998853$, which corresponds to an angle of $177.3^{\circ}$ or $3.09$ radians.

\begin{equation}
G2_{00}  = \frac{ 0.2950 + 2\sqrt{0.17}\sqrt{0.125}\cos(3.09) }{ 0.5 + 0.4915\cos(3.09) } = 0.42 
\label{eq:g00}
\end{equation}

\begin{equation}
G2_{11} = \frac{ 0.2050 + 2\sqrt{0.08}\sqrt{0.125}\cos(3.09) } { 0.5 + 0.4915\cos(3.09) } = 0.58
\label{eq_g11_final}
\end{equation}

This section described the application of the proposed quantum interference Bayesian Network to explain the puzzling findings in the two-stage gambling game, that could not be explained through classical probability theory. Equations~\ref{eq:g00} and~\ref{eq_g11_final} showed that the proposed approach is able to simulate the human decisions observed in the works of~\cite{Tversky92,Shafir92,Kuhberger01,Lambdin07}. 

\subsection{The Impact of the Phase $\theta$}

The interference term that we show in this paper for the two-stage gambling game, was proposed by cognitive psychologists in order to explain their observations. That is, they manually tuned this parameter $\theta$ to fit their data. In this work, we look at this parameter from a different perspective: what happens to the quantum computed probabilities if we vary this angle $\theta$? Can better inferences be achieved? In this section, we make use of the probabilistic interference term proposed in the cognitive psychology literature~\cite{Busemeyer09,busemeyer06}, and investigate the impact that the phase parameter $\theta$ can have,when computing quantum probabilities in the two step gambling game.

\begin{figure}[h!]
\resizebox{\columnwidth}{!} {
	\includegraphics{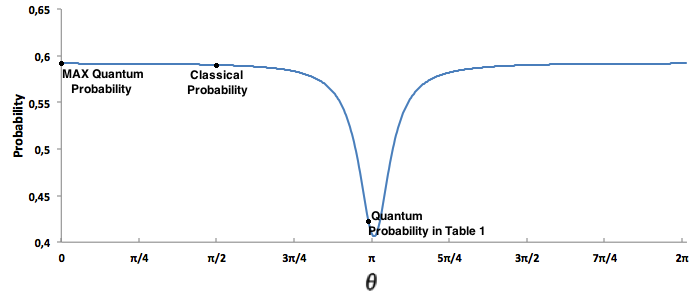}
	}
	\caption{The various quantum probability values that can be achieved by variying the angle $\theta$ in Equation~\ref{eq:g00_norm}.  Note that quantum probability can achieve much higher/lower values than the classical probability.}
	\label{fig:impact_theta_gambling}
\end{figure}

In the previous section, the general probability formula of a player willing to play the second gamble, given that the outcome of the first gamble was unknown, was given by Equation~\ref{eq:g00_norm}. In order to analyze the consequences of the angle $\theta$ in the final probability, we varied $\theta$ from $0$ to $2\pi$ in  steps of $0.0001$ radians. The results obtained are discriminated in Figure~\ref{fig:impact_theta_gambling}.

Figure~\ref{fig:impact_theta_gambling} reveals that quantum probabilities can achieve much higher values than the classical probability theory. The quantum probability can reach a maximum of $0.5915$ where the classical probability can only have a fixed value of $0.59$. 

\begin{figure}[ht]
	\parbox{.45\linewidth}{
	\centering
	\includegraphics[scale=0.3]{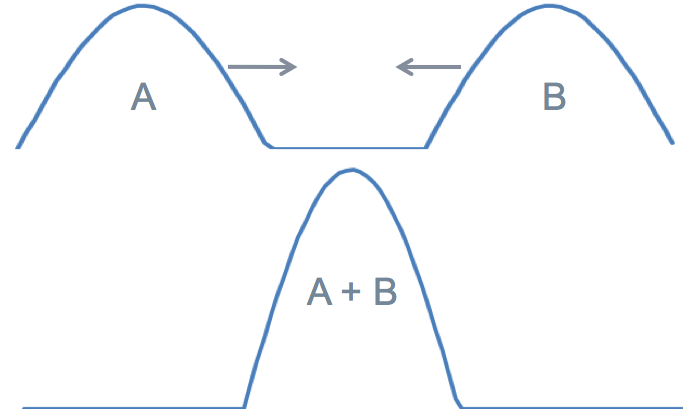}
	\caption{Example of constructive interference: two waves collide forming a bigger wave.}
	\label{fig:const_inter}
	}
	\hfill
	\parbox{.45\linewidth}{
	\centering
	\includegraphics[scale=0.3]{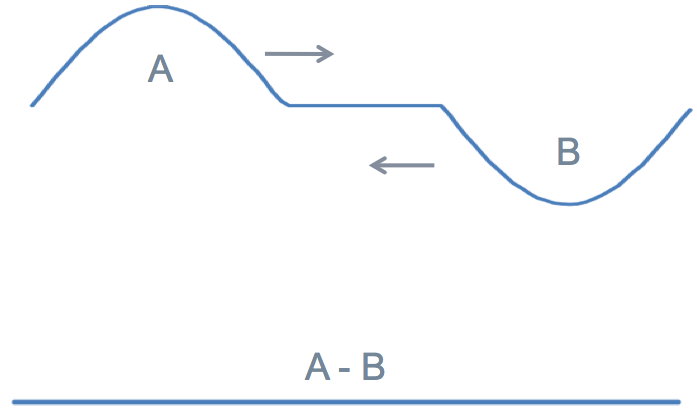}
	\caption{Example of destructive interference: two waves collide cancelling each other.}
	\label{fig:destruct_inter}	
	}
\end{figure}

Figure~\ref{fig:impact_theta_gambling} is also supporting the quantum information processing theory already mentioned in Section~\ref{sec:intro} of this work: information is modeled via wave functions and therefore, they cannot be in a definite state (only when a final decision is made, a definite state emerges). One can look at all values that this parameter $\theta$ as all possible probabilities (or outcomes) that a player has when deciding to whether or not to play the second gamble. Since in quantum theory we model the participant's beliefs by wave functions, then the superposed states can produce different waves coming from opposite directions that can crash into each other. When they crash, the waves can either unite or be destroyed. When they unite, it causes a constructive interference effect that will cause a bigger wave, leading to a maximum or minimum quantum probability value, depending on the phase of the wave (Figure~\ref{fig:const_inter}). When the waves crash and are destroyed, then a destructive interference effect occurs (Figure~\ref{fig:destruct_inter}).

In conclusion, quantum probability enables the free choice of parameters in order to obtain a desired probability value. The two-stage gambling game is just a small example where the proposed model could be applied. In the next section, we turn to the task of burglary detection. We will analyse a more complex Bayesian Network that will determine the probability of a burglary occurring, given that the neighbours think that they heard an alarm.

\section{Inference in More Complex Networks: The Burglar/Alarm Network}\label{sec:classic_vs_quantum_burglar}

In this section, we compare classical inferences in Bayesian Networks with the proposed quantum model. The example that we will examine correspond to a modified version of the Burglar/Alarm network, which is very used in the scope of Artificial Intelligence~\cite{Pearl88,russel10}. 

The network proposed in the book of~\cite{russel10} is inspired in the following scenario. Imagine that a person installed a new burglar alarm in his house. The alarm has a big credibility in what concerns detecting burglaries. The person has two neighbours: Mary and John. They have promised to call the person, whenever they think they heard the alarm. John always calls the police, when he hears the alarm, however, he sometimes confuses the sound of the alarm with the ringtone of his phone, resulting in a misleading phone call to the police. Mary, on the other hand, cannot hear the alarm very often, because she likes to hear very loud music. Given the evidence of who has or has not called the police, we want to represent in a Bayesian Network the probability of a burglary occurring~\cite{russel10}. 

\begin{figure}[ht]
	\parbox{.45\linewidth}{
	\centering
	\includegraphics[scale=0.36]{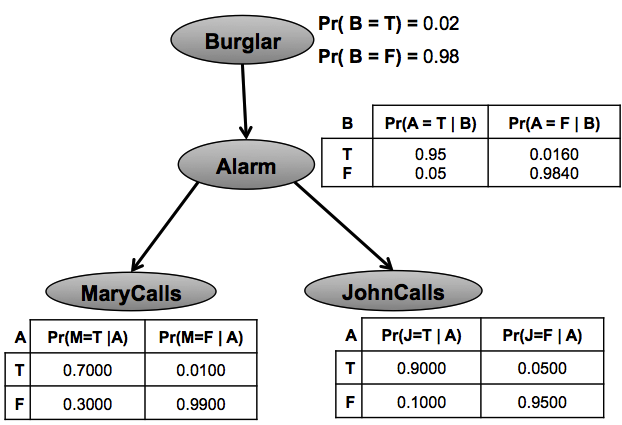}
	\caption{Burglar/Alarm classical Bayesian Network proposed in the book of~\cite{russel10}}
	\label{fig:earthquake_bn}
	}
	\hfill
	\parbox{.5\linewidth}{
	\centering
	\includegraphics[scale=0.36]{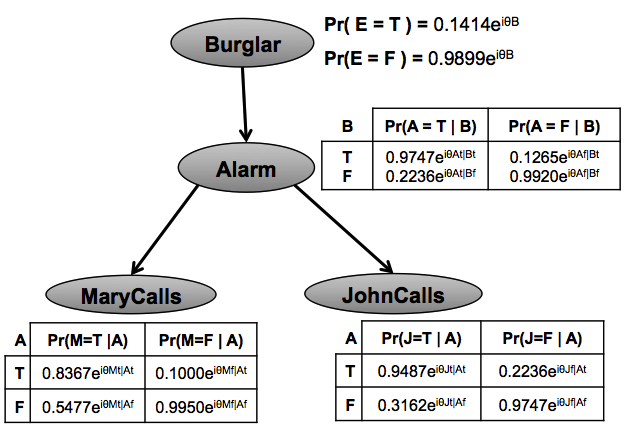}
	\caption{ Quantum counterpart of the Burglar/Alarm Bayesian Network proposed in the book of~\cite{russel10}}
	\label{fig:earthquake_qbn}	
	}
\end{figure}

Figure~\ref{fig:earthquake_bn} represents a classical Bayesian Network to account for burglary detection. It's quantum counterpart corresponds to Figure~\ref{fig:earthquake_qbn}. In order to make a fair comparison between both networks, the quantum Bayesian Network was built in the same way as the classical one, but we replaced the real probability values by quantum complex amplitudes, just like proposed in~\cite{Tucci95} and~\cite{Leifer08}. 

We also performed a set of queries so we could compare the classical Bayesian Network with the proposed quantum interference Bayesian Network. Table~\ref{tab:classical_results} presents the final probabilities computed over the classical Bayesian Network of Figure~\ref{fig:earthquake_bn} and Table~\ref{tab:earthquake_qbn_results} presents the results for the quantum Bayesian network (Figure~\ref{fig:earthquake_qbn}). 

\begin{table}
\resizebox{\columnwidth}{!} {
\begin{tabular}{l | c | c | c | c | c |}
	\bf{Evidences}			& \bf{Pr( Alarm = t )}	 & \bf{Pr( Burglar = t )}	& \bf{Pr( JohnCalls = t )}	& \bf{Pr( MaryCalls = t )} \\
			\hline		
\bf{No Evidence} 			& 0.0347 				& 0.0200 				& 0.0795 				& 0.0339	\\
\bf{Alarm = t}				& 1.0000				& 0.5479 				& 0.9000 				& 0.7000	\\
\bf{Burglar = t}				& 0.9500				& 1.0000 				& 0.6655 				& 0.8575	\\	
\bf{JohnCalls = t}			& 0.3927 				& 0.2158 				& 1.0000 				& 0.2810	\\	
\bf{MaryCalls = t}			& 0.7155 				& 0.3923 				& 0.6582 				& 1.0000	\\	
\hline	
\hline	
\bf{Alarm = t, Burglar = t} 		& 1.0000 		& 1.0000 				& 0.7000 				& 0.9000	\\
\bf{Alarm = t, JohnCalls = t} 		& 1.0000 		& 0.5479 				& 1.0000 				& 0.7000	\\
\bf{Alarm = t, MaryCalls = t} 		& 1.0000 		& 0.5479 				& 0.9000 				& 1.0000	\\
\bf{Burglar = t, JohnCalls = t} 		& 0.9971 		& 1.0000 				& 1.0000 				& 0.6980	\\
\bf{Burglar = t, MaryCalls = t} 		& 0.9992 		& 1.0000 				& 0.8994 				& 1.0000	\\
\bf{JohnCalls = t, MaryCalls = t} 	& 0.9784 		& 0.5360 				& 1.0000 				& 1.0000	\\
\hline	
\end{tabular}
}
\caption{Probabilities obtained when performing inference on the classical Bayesian Network of Figure~\ref{fig:earthquake_bn}.}
\label{tab:classical_results}
\end{table}

\begin{table}[h!]
\resizebox{\columnwidth}{!} {
\begin{tabular}{l | c | c | c | c | c |}
	\bf{Evidences}		& \bf{Pr( Alarm = t )} 	& \bf{Pr( Burglar = t )} 	& \bf{Pr( JohnCalls = t )} 		& \bf{Pr( MaryCalls = t )} \\
			\hline		
\bf{No Evidence} 		& 0.1760 				& 0.1179 				& 0.1185 				& 0.0889 	\\
\bf{Alarm = t}			& 1.0000				& 0.5479 				& 0.9000 				& 0.7000	\\
\bf{Burglar = t}			& 0.9896				& 1.0000 				& 0.9999 				& 0.9791	\\	
\bf{JohnCalls = t}		& 0.6596 				& 0.8380 				& 1.0000 				& 0.8138	\\	
\bf{MaryCalls = t}		& 0.9018 				& 0.9998 				& 0.9758 				& 1.0000	\\	
\hline	
\hline	
\bf{Alarm = t, Burglar = t} 		& 1.0000 		& 1.0000 				& 0.9000 				& 0.7000	\\
\bf{Alarm = t, JohnCalls = t} 		& 1.0000 		& 0.5479 				& 1.0000 				& 0.7000	\\
\bf{Alarm = t, MaryCalls = t} 		& 1.0000 		& 0.5479 				& 0.9000 				& 1.0000	\\
\bf{Burglar = t, JohnCalls = t} 		& 0.9982 		& 1.0000 				& 1.0000 				& 0.7390	\\
\bf{Burglar = t, MaryCalls = t} 		& 0.9993 		& 1.0000 				& 0.9138 				& 1.0000	\\
\bf{JohnCalls = t, MaryCalls = t} 	& 0.9883 		& 0.6632 				& 1.0000 				& 1.0000	\\
\hline		
\end{tabular}
}
\caption{Probabilities obtained when performing inference on the quantum Bayesian Network of Figure~\ref{fig:earthquake_qbn}.}
\label{tab:earthquake_qbn_results}
\end{table}

In the experiment of the two-stage gambling game, we saw that the parameter $\theta$ plays an important role in determining the final probability value of a variable. For the Burglar/Alarm Bayesian Network, in order to find the best parameter for each query, we varied $\theta$ between $0$ and $2\pi$, in steps of $0.1$ radians, and collected the $\theta$ that would maximize most variables in the network. Equation~\ref{eq:max_interf} represents the interference formula that we used to maximize the probability with respect to parameter $\theta$. 

\begin{equation}
Pr(A)= argmax_{\theta} ~~ \alpha \left[ \sum_{i=1}^N | \psi_{i} |^2 + 2 \sum_{i=1}^{N-1} \sum_{j=i+1}^{N} | \psi_i | | \psi_j | \cos( \theta_i - \theta_j ) \right]
\label{eq:max_interf}
\end{equation}

In the next section, we will analyse the results specified in Tables~\ref{tab:classical_results} and~\ref{tab:earthquake_qbn_results} for each single query.

\section{Discussion of Experimental Results}\label{sec:discussion}

Tables~\ref{tab:classical_results} and~\ref{tab:earthquake_qbn_results} present the results obtained for different queries performed over the classical Bayesian Network (Figure~\ref{fig:earthquake_bn}) and the proposed quantum interference Bayesian Network (Figure~\ref{fig:earthquake_qbn}), respectively.

Analysing the first row of both tables, when we provide no piece of evidence to the network, the proposed model was able to increase, on average, the probabilities of the query variables about $270.625\%$. In the case where nothing is observed, the proposed network achieves its maximum level of uncertainty: all variables are interfering with each other causing both destructive and constructive interference effects on the final probabilities. It is interesting to notice that when the classical probabilities computed for each query variable are very low, then the quantum Bayesian Network cannot greatly increase these probabilities. If the probabilities in a classical setting are very low, then the quantum Network is able to keep those probabilities low as well.

\begin{figure}[h!]
	\parbox{.45\linewidth}{
	\centering
	\includegraphics[scale=0.2]{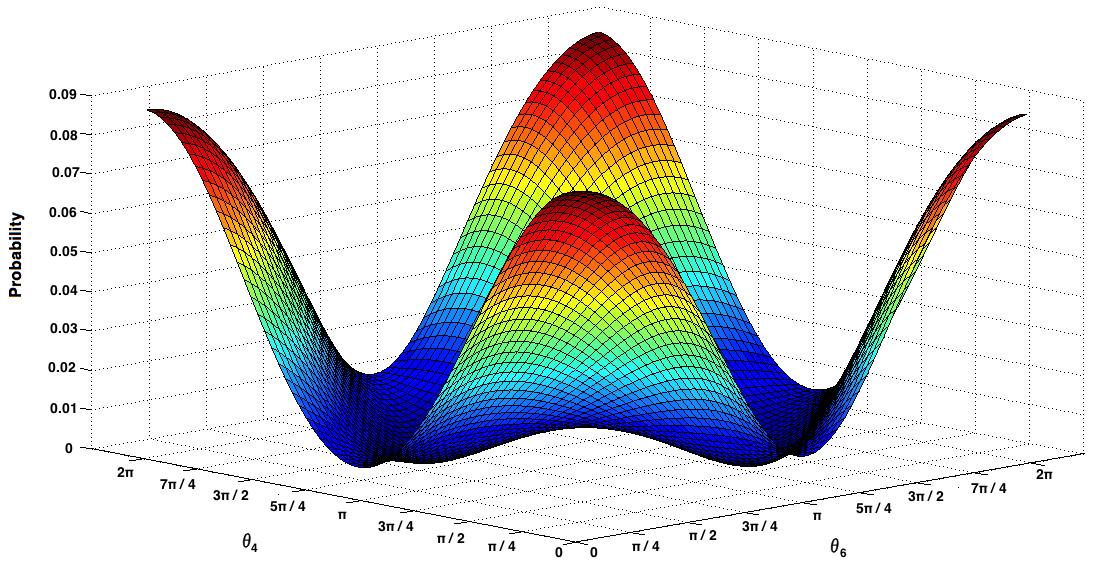}
	\caption{Possible probabilities when querying "MaryCalls = t" with no evidence. Parameters used were: $\{\theta_1,\theta_2,\theta_3,\theta_5,\theta_7,\theta_8 \}\rightarrow \{0,0,0,0,3.1,0  \}.$ Maximum probability for $\{\theta_1,\theta_2 \}\rightarrow \{0,3.1 \}.$}
	\label{fig:marycalls}
	}
	\hfill
	\parbox{.45\linewidth}{
	\centering
	\includegraphics[scale=0.2]{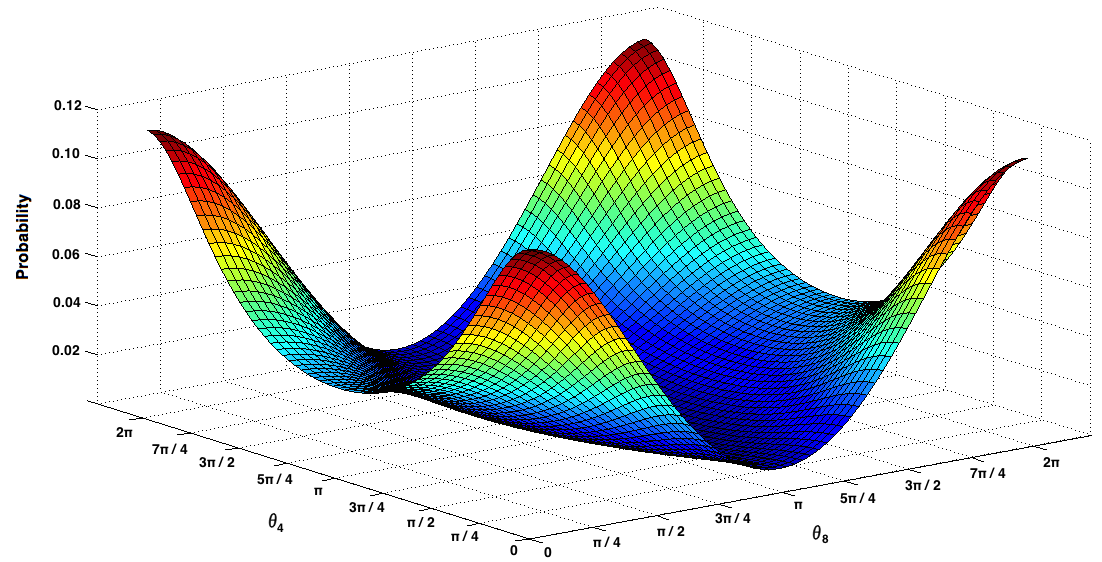}
	\caption{Possible probabilities when querying "Burglar = t" with no evidence. Parameters used were: $\{\theta_1,\theta_2,\theta_3,\theta_5,\theta_6,\theta_7 \}\rightarrow \{0,0,0,6.2,0.1,3.1 \}.$  Maximum probability for $\{\theta_4,\theta_8 \}\rightarrow \{0,3.2 \}.$}
	\label{fig:burglar}
	}
\end{figure}

\begin{figure}[h!]
	\parbox{.45\linewidth}{
	\centering
	\includegraphics[scale=0.2]{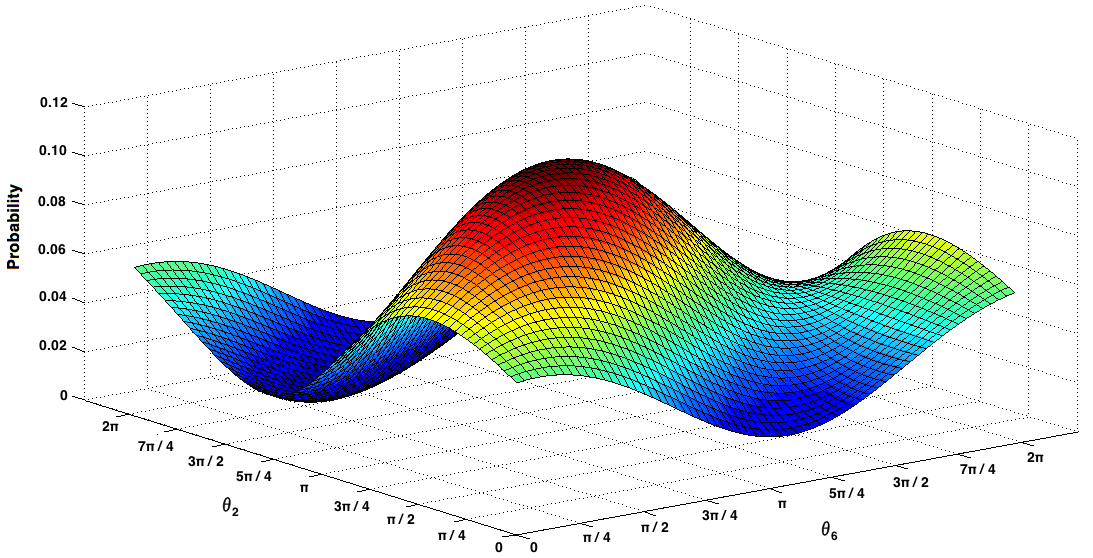}
	\caption{Possible probabilities when querying "JohnCalls = t" with no evidence. Parameters used were: $\{\theta_1,\theta_3,\theta_4,\theta_5,\theta_7,\theta_8 \}\rightarrow \{1.9,0,2.3,0.5,4.5,2.4  \}.$  Maximum probability for $\{\theta_2,\theta_6 \}\rightarrow \{2.3,5.5 \}.$}
	\label{fig:johncalls}
	}
	\hfill
	\parbox{.45\linewidth}{
	\centering
	\includegraphics[scale=0.2]{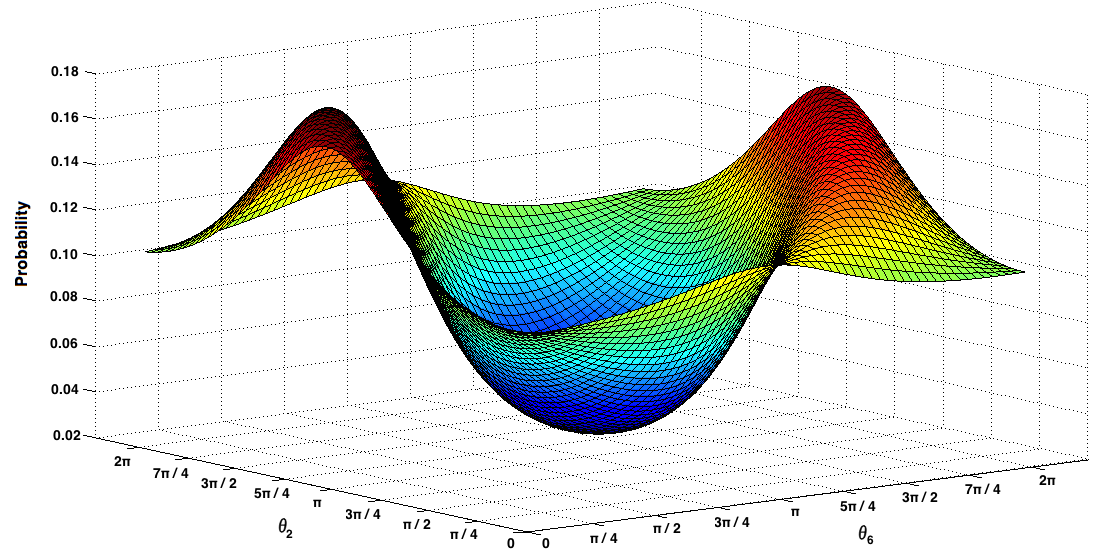}
	\caption{Possible probabilities when querying "Alarm = t" with no evidence. Parameters used were: $\{\theta_1,\theta_3,\theta_4,\theta_5,\theta_7,\theta_8\}\rightarrow \{0,0,0.8,6.2,3.1,4.3\}.$  Maximum probability for $\{\theta_2,\theta_6 \}\rightarrow \{0.2,0.5 \}.$}
	\label{fig:alarm}
	}
\end{figure}

Figures~\ref{fig:marycalls}-\ref{fig:alarm} illustrate all possible values that each query variable could have, by varying the $\theta$ parameters. These graphs were plotted in the following way: for each query, we found the set of all $\theta$'s that would lead to a maximum and  a minimum probability value. We then compared the set of $\theta$'s and realised that in the majority of the cases there were components that shared the same parameters. Given this situation, we fixed 6 of the parameters which were common and varied the remaining parameters, leading to a 3-dimensional graph. 


The moment we start to provide pieces of evidence to the network, the uncertainty starts to decrease. Analyzing the situation where we observe that the $Alarm$ variable is true, then an interesting phenomena occurs: the probabilities of the proposed quantum Bayesian Network collapse to the same probability values as in its classical counterpart. If one observes the variable $Alarm$, then the variables $Burglar$, $MaryCalls$ and $JohnCalls$ become independent of each other. This means that there is no possible way that these variables can interfere with each other. Since the variables do not provoke any interferences among them, then the interference term will be zero and will collapse to the classical probability. This independence phenomena in quantum Bayesian Networks has also been noticed in the work of~\cite{Leifer08}. For a mathematical proof of why this phenomena also occurs in a quantum setting, please refer to their work. This means that, whenever we observe the variable $Alarm$, the interference term will always be null and consequently all probabilities will be exactly the same as in a classical Bayesian Network inference.

In the case where we observe the $Burglar$ variable, then, according to the scenario of the Bayesian Network, the $Alarm$ variable should also increase. The variable $JohnCalls$ is highly correlated with the variable $Alarm$. In its conditional probability table, there is a chance of $90\%$ of the variable $JohnCalls$ occurring when the variable $Alarm$ is true. So in this situation, the quantum Bayesian Network is able to give more strength to that correlation. On average, the probabilities of all queries increased $22.8\%$ when the variable $Burglar$ is observed, when compared to the respective classical setting.

When we observe that the variable $JohnCalls$ is true, then it is expected that the probability of the $Alarm$ variable increases as well, according to the scenario of the Bayesian Network. When we observe the variable $MaryCalls$ the variable $Alarm$ increases even more when compared to the situation where $JohnCalls~=~t$ or its classical counterpart. This means that the correlation between the $Alarm$ variable and the variable $JohnCalls$ is not as strong as when we observe $MaryCalls$. According to the scenario, John always calls the police, leading to many misleading calls. Consequently, the $Alarm$ variable cannot be increased too much. However, when $MaryCalls$ is observed, then, according to the Bayesian scenario, it is almost certain that she heard the alarm and therefore, a strict correlation exists between these two variables. The quantum Bayesian Network was able to represent the correlations between variables in a more realistic and reliable way than its classical counterpart. 

Finally, when we start to provide 2 pieces of evidence  to the network, then the uncertainty levels start to decrease. Consequently, the probabilities computed in a quantum setting start to converge to the ones computed in a classical Bayesian Network. This means that, there are two situations there the proposed quantum Bayesian Network converges to its classical counterpart: (1) when the variables of the network become independent of each other and (2) when there are very low levels of uncertainty, because too many evidences were provided to the network.

\section{The Optimum Value for $\theta$ to Maximize Quantum Inferences in Bayesian Networks }\label{sec:opt}

\begin{figure}[ht]
	\parbox{.45\linewidth}{
	\centering
	\includegraphics[scale=0.3]{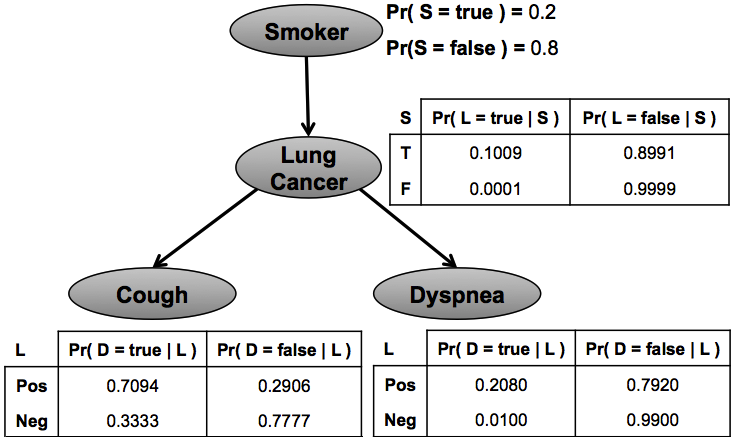}
	\caption{Classical representation of a Lung Cancer Bayesian Network inspired in the book of~\cite{Pearl88}.}
	\label{fig:lung_cancer_bn}
	}
	\hfill
	\parbox{.5\linewidth}{
	\centering
	\includegraphics[scale=0.3]{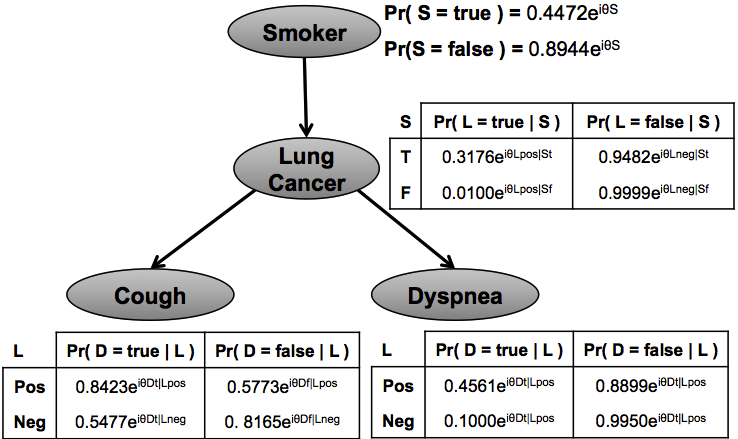}
	\caption{ Quantum representation of a Lung Cancer Bayesian Network inspired in the book of~\cite{Pearl88}.}
	\label{fig:lung_cancer_qbn}	
	}
\end{figure}

In this section, we perform a study on the impact of the $\theta$ parameters in the context of medical decision making. Consider the Classical Bayesian Network of Figure~\ref{fig:lung_cancer_bn}, which corresponds to a slightly modified version of the Bayesian Network proposed in the book of~\cite{Pearl88}. Figure~\ref{fig:lung_cancer_qbn} corresponds to its quantum counterpart. 

\begin{figure}[h!]
\centering
\includegraphics[scale=0.5]{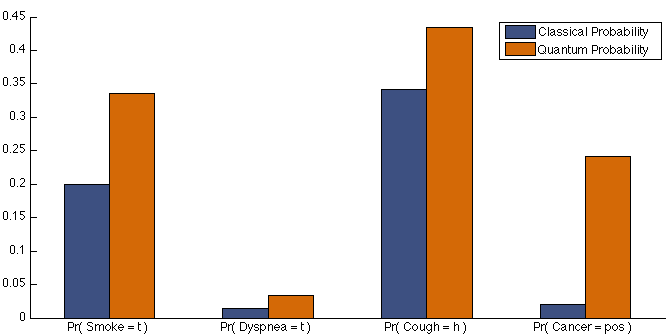}
\caption{Classical and Quantum probabilities computed for different queries for the Lung Cancer Bayesian Network in Figure~\ref{fig:lung_cancer_bn}.}
\label{fig:cancer_results}
\end{figure}

Again, in order to determine the maximum probability value, we varied the $\theta$ parameters between 0 and $2\pi$ in steps of $0.1$ radians. The method that we used to compute these parameters in described in Appendix~\ref{sec:grid}. We then performed the following queries: Pr( Smoke = true ), Pr( Dyspnea = true ), Pr( Cough = high ) and Pr( Lung Cancer = positive ). The results obtained are discriminated in Figure~\ref{fig:cancer_results}.

Analysing Figure~\ref{fig:cancer_results}, one can observe that again, the quantum probability values tend to overcome their classical counterparts. A special note goes to the quantum probabilities verified for the query Pr( Cancer = positive ). The increase verified in this probability was much higher than any other value achieved by the other quantum probabilities. Since the $Lung\_Cancer$ random variable is located at the centered position of the Bayesian Network, then, when nothing is observed, its probability is influenced by the probabilities of all nodes in the network. 

In order to determine the impact of the $\theta$ parameters, Table~\ref{tab:quantum_results_lung} shows the parameters that were used in the Burglar/Alarm and  Lung Cancer Bayesian Networks.

\begin{table}[h!]
\centering
\begin{tabular}{l | c | c | c | c | c | c | c | c }
	\bf{Variables}			& $\theta_1$	& $\theta_2$	& $\theta_3$	& $\theta_4$	& $\theta_5$	& $\theta_6$	& $\theta_7$	& $\theta_8$ 	 \\
			\hline		
\bf{Alarm = t} 				& 0.00 		& 0.20 		& 0.00 		& 0.80		& 6.20 		& 0.50 		& 3.10 		& 4.30		\\
\bf{Burglar = t}				& 0.00 		& 0.00 		& 0.00 		& 0.00		& 6.20 		& 0.10 		& 3.10 		& 3.20		\\
\bf{JohnCalls = t} 			& 1.90 		& 2.30 		& 0.00 		& 2.30		& 0.50 		& 5.50 		& 4.50 		& 2.40		\\
\bf{MaryCalls = t} 			& 0.00 		& 0.00 		& 0.00 		& 0.00		& 0.00 		& 3.10 		& 3.10 		& 0.00		\\
\hline	
\hline
\bf{Cancer = t} 			& 3.20 		& 3.20 		& 3.20 		& 3.20		& 3.20 		& 3.20 		& 0.10 		& 0.00		\\
\bf{Cough = t} 				& 0.00 		& 0.00 		& 0.00 		& 0.00		& 3.10 		& 3.20 		& 0.10 		& 0.00		\\
\bf{Dyspnea = t} 			& 0.00 		& 0.00 		& 0.00 		& 0.00		& 3.20 		& 3.20 		& 0.10 		& 0.00		\\
\bf{Smoke = t} 				& 0.00 		& 0.00 		& 0.00 		& 0.00		& 3.20 		& 3.20 		& 0.10 		& 0.00		\\
\hline
\end{tabular}
\caption{Optimum $\theta$'s found for each variable from the burglar/alarm bayesian network (Figure~\ref{fig:earthquake_bn}) and from the lung cancer bayesian network (Figure~\ref{fig:lung_cancer_bn}).}
\label{tab:quantum_results_lung}
\end{table}

\begin{figure}[h!]
	\parbox{.45\linewidth}{
	\centering
	\includegraphics[scale=0.2]{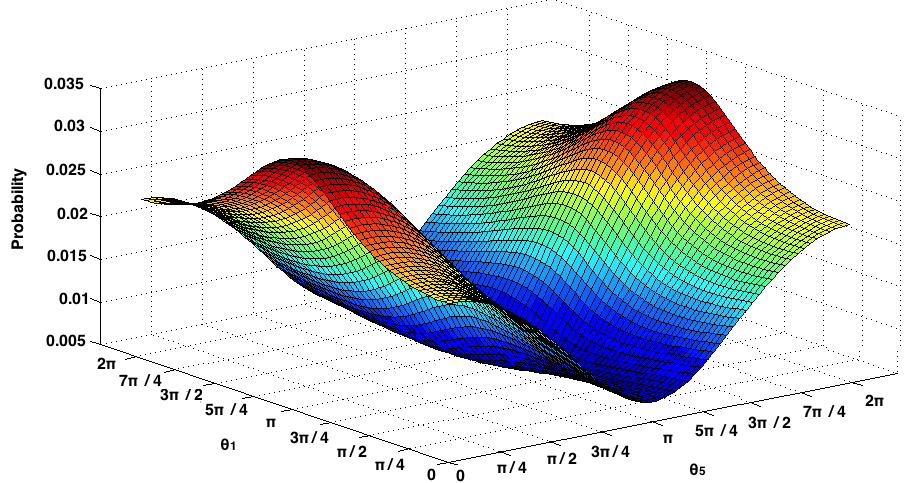}
	\caption{Possible probabilities when querying "Dyspnea = t" with no evidence. Parameters used were: $\{\theta_2,\theta_3,\theta_4,\theta_6,\theta_7,\theta_8 \}\rightarrow \{0,0,0,3.2,0.1,0  \}.$ Maximum probability for $\{\theta_1,\theta_5 \}\rightarrow \{0,3.2 \}.$}
	\label{fig:dyspnea}
	}
	\hfill
	\parbox{.45\linewidth}{
	\centering
	\includegraphics[scale=0.2]{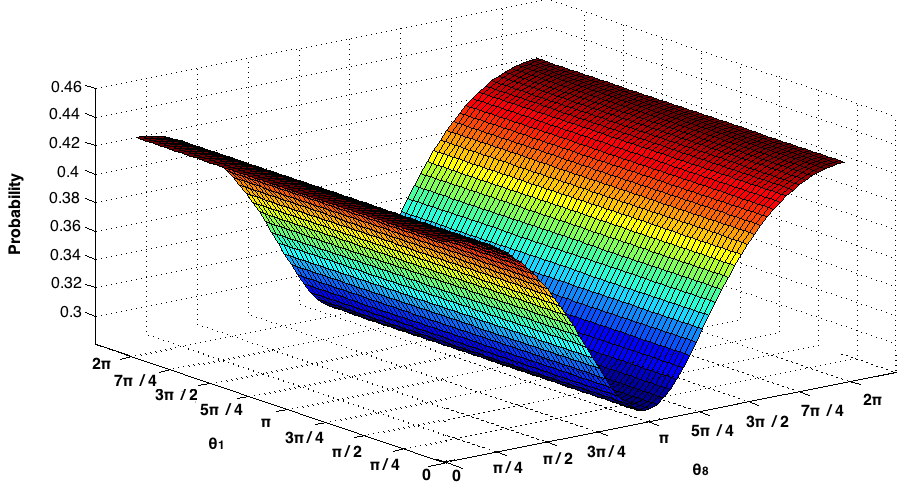}
	\caption{Possible probabilities when querying "Cough = t" with no evidence. Parameters used were: $\{\theta_2,\theta_3,\theta_4,\theta_5,\theta_6,\theta_7 \}\rightarrow \{0,0,0,3.1,3.2,0.1 \}.$  Maximum probability for $\{\theta_1,\theta_8 \}\rightarrow \{0,3.1 \}.$}
	\label{fig:cough.png}
	}
\end{figure}

\begin{figure}[h!]
	\parbox{.45\linewidth}{
	\centering
	\includegraphics[scale=0.2]{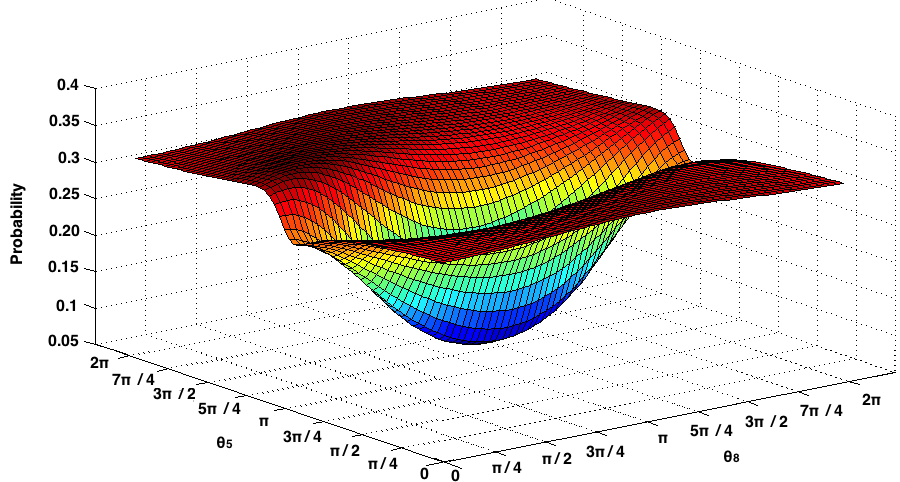}
	\caption{Possible probabilities when querying "Smoke = t" with no evidence. Parameters used were: $\{\theta_1,\theta_2,\theta_3,\theta_4,\theta_6,\theta_7 \}\rightarrow \{0,0,0,0,0.1,0  \}.$  Maximum probability for $\{\theta_5,\theta_8 \}\rightarrow \{0,0.8 \}.$}
	\label{fig:smoke}
	}
	\hfill
	\parbox{.45\linewidth}{
	\centering
	\includegraphics[scale=0.2]{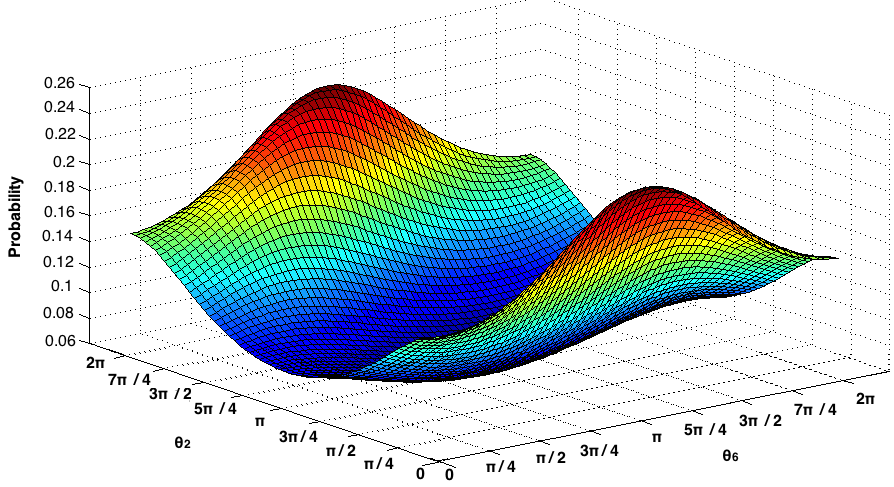}
	\caption{Possible probabilities when querying "Cancer = pos" with no evidence. Parameters used were: $\{\theta_1,\theta_2,\theta_3,\theta_4,\theta_5,\theta_7\}\rightarrow \{3.2,3.2,3.2,3.2,3.2,0.1\}.$  Maximum probability for $\{\theta_2,\theta_6 \}\rightarrow \{3.2,3.2 \}.$}
	\label{fig:cancer}
	}
\end{figure}

Figures~\ref{fig:dyspnea}-\ref{fig:cancer} also show all the possible probability values that the variables from the lung cancer Bayesian network can achieve, when nothing is known.

When we are at a maximum level of uncertainty, classical probabilities tend to assume that events are equiprobable, that is, the probability of their outcome is always the same no matter their context. Quantum theory, on the other hand, provides a more relaxed framework. When nothing is known, then there are more degrees of freedom that enable the outcome of any event to be any possible value. This way, by analysing the context of certain events (for example, a medical decision scenario, a gambling game, etc), the quantum parameters $\theta$ can be modelled in such a way that they can simulate reality more accurately and more precisely than its classical counterpart. 

In this sense, in order to develop more accurate quantum Bayesian Networks, a study of the context of the problem is required in order to start the search for the optimum $\theta$ parameters. Techniques to search for these optimum quantum parameters is still an open research question and an unexplored field in the literature of quantum cognition and quantum decision models.

\section{Related Work}\label{sec:rel_work}

Since the preliminary findings of~\cite{Tversky74}, cognitive scientists began to use alternative theories, such as quantum probability theory, in an attempt to explain the paradoxical findings that classical probability theory could not explain. Given the flexibility of the quantum probabilistic framework, several researchers applied these concepts in several fields outside of physics. In this section, we present some of the topics in which quantum probability has made some impacts in the literature.

\subsection{Violations of Probability Theory}

In what concerns violations of probability theory, there are many paradoxical situations where classical probability is unable to model. The most important paradoxes consist in violations in the Sure Thing principle. Examples of such violations correspond to the Ellsberg paradox~\cite{Ellsberg61}, the two-stage gambling game~\cite{Tversky92,Kuhberger01,Lambdin07} and the prisoner's dilemma game~\cite{Tversky92,Croson99,Li02,Busemeyer06proceed,Hristova08,Conte07experiment}.
	
In what concerns the prisoner's dilemma game, many works in the literature have been proposed, which  formalise the problem in a quantum approach. For instance,~\cite{Busemeyer09} proposed a quantum dynamic Markov Model for the prisoner's dilemma game. In their model, the authors represented the players beliefs and actions as a superposition in a quantum state vector. When the player knew the information about the opponent's action, the superposition state collapsed to a new state that was compatible with the action chosen by the opponent. When the player did not know about the opponent's move, then the system involved through the application of unitary transformations. In~\cite{Asano10}, the authors focus on irrational choices and developed a model based on the quantum superposition and interference principles for the prisoner dilemma game. After each play, the state is updated until it reaches an equilibrium. In~\cite{Cheon10}, the authors analysed quantum conditional probabilities in the prisoner's dilemma game. There are also works that consider that the probabilities in the prisoner's dilemma game correspond to different contexts and these contexts can be incompatible. The authors argue that the probabilities observed in the prisoner's dilemma game were not classical, but were not quantum either in the sense of traditional quantum mechanics theory. So the author introduced the quantum-like approach and suggested a new trigonometric interference term~\cite{Khrennikov09quantumlike,Khrennikov09sure}.~\cite{Yukalov11} proposed quantum model which incorporated entangled decisions and explained the violations of the Sure Thing principle through interference effects between intentions. In~\cite{Accardi09}, the authors explored the impact of these violations in economics and analyse the two-stage gambling game and the prisoner's dilemma game under a quantum probabilistic point of view, by proposing a quantum Markov Model to explain the paradoxical observations in these games. Finally,~\cite{Asano12} proposed a quantum bayesian updating scheme based on the formalisms of quantum mechanics that represents mental states on a Hilbert state. Through the usage of projections, they were able to introduce a model that could explain the paradoxical findings of the two-stage gambling game~\cite{Tversky92}. Another similar work that compares Bayes rule to its quantum counterpart corresponds to~\cite{Busemeyer09comparison}.

\subsection{Conjunction and Diskunction Errors}

Other situations, in which the laws of probability are being violated, correspond to disjunction or conjunction fallacies. A conjunction error occurs when it is assumed that specific conditions are more probable than a single general one. A disjunction error, on the other hand, consists in assuming that the disjunction of two events is at least as likely as either of the events individually. Several works in the literature used quantum probabilistic models to address these fallacies. For instance,~\cite{Franco09} developed a quantum probabilistic model to explain fallacies when making preferences over a set of events, more specifically, the conjunction fallacy.~\cite{Busemeyer11} also focused on quantum models to explain conjunction and~\cite{Croson99} focused on disjunction fallacies. The first experiments where these fallacies were observed to occur were performed by~\cite{Tversky92}. An alternative model for conjunction and disjunction errors correspond to~\cite{Tversky94}. The authors used support theory as a subjective probabilistic model that describes the way people make probability judgments. The main problem of their model is that support theory is able to successfully describe conjunction errors, but fails at explaining disjunction errors. Other models correspond to~\cite{Gigerenzer95,Gigerenzer96}.~\cite{Yukalov11} also addressed the problem of conjunction and disjunction fallacies. They proposed quantum model which incorporated entangled decisions.~\cite{Khrennikov06} also modelled mental processes through quantum probabilities, where the interference process plays an important role in the process of recognising images\cite{Conte09}. Other interesting works of this author applying similar quantum formalisms correspond to~\cite{Khrennikov09quantumlike,Khrennikov09,Khrennikov07,Tentori13}.

\subsection{Quantum Probabilistic Graphical Models}

In what concerns quantum probabilistic graphical models, there are various contributions. For example, in~\cite{Tucci95} the model proposed by the author is exactly the same as a classical Bayesian Network. The only difference is that it uses complex numbers in the conditional probability tables instead of real values. A similar model has been proposed by~\cite{Mura07}, but for Markov Networks. In~\cite{Leifer08}, the author proposed a quantum Bayesian Network by replacing the classical formulas used to perform the inference process by their quantum counterpart. \cite{busemeyer06},~\cite{Busemeyer06proceed} and~\cite{Busemeyer09markov} proposed a quantum dynamic Markov model based on the findings of cognitive psychologists and interference terms. 


\subsection{General Applications of Quantum Probabilities}

 One of the first and most influential models that applied quantum probability for decision making belongs to~\cite{Aerts09}. In their work, the authors proposed the $\epsilon$-model, which can be defined as a quantum machine that corresponds to a two dimensional  Hilbert space. Given an event in this quantum machine one can compute quantum like probabilities. The model also makes use of a parameter $\epsilon$ that measures the total amount of uncertainty when computing the probabilities. Other works related to this model can be found in~\cite{Aerts93physical,Aerts93quantum_probability,Aerts94quantum,Aerts95quantum_structures,Aerts05a,Aerts05b,Aerts95quantum_structures,Aerts96,Aerts98,Gabora02,Aerts11}.

\cite{Asano12} proposed a quantum bayesian updating scheme based on the formalisms of quantum mechanics that represents mental states on a Hilbert state. Through the usage of projections, they were able to introduce a model that could explain the paradoxical findings of the two-stage gambling game~\cite{Tversky92}. Another similar work that compares Bayes rule to its quantum counterpart corresponds to~\cite{Busemeyer09comparison}.

There is also an increasing interest in applying quantum like models to subjects outside of psychology. In game theory there have been works exploring the uncertainties of a player towards another through quantum probabilistic models. For instance, \cite{Asano10} focus on irrational choices and developed a model based on the quantum superposition and interference principles for the Prisoner Dilemma game. After each play, the state is updated until it reaches an equilibrium.~\cite{Cheon10} used the data collected by~\cite{Tversky92} and analysed quantum conditional probabilities.~\cite{Mura09} generalised the classical expected utility and proposed a quantum projective expected utility function that does not violate the laws of classical probability theory and can accommodate Allais and Ellsberg paradoxes.~\cite{Danilov08} made a formalisation of the structure of quantum probability theory and~\cite{Danilov10} also applied these formalists to expected utilities.~\cite{Piotrowski01} applied quantum theory to the fields of economics and game theory and developed a principle to minimise financial risks ~\cite{Aerts04} modelled how a person updates its beliefs in the liar's paradox.~\cite{Khrennikov06} also modelled mental processes through quantum probabilities, where the interference process plays an important role in the process of recognising images\cite{Conte09}. Other interesting works of this author applying similar quantum formalisms correspond to~\cite{Khrennikov09quantumlike,Khrennikov09sure,Khrennikov09,Khrennikov07,Khrennikov99,Haven13}.~\cite{Mogiliansky09} developed a quantum mechanics based framework in order to model agents preferences under uncertainty.
	
More recently,~\cite{Busemeyer12quantum} published a work that generated discussions over the scientific literature, about whether or not, quantum probability can provide a new direction to compute quantum probabilistic inferences. In their work, they summarised the main areas where quantum probability has made some impact (conjunction/disjunction errors, sure thing principle, etc). A deep analysis of the differences between classical and quantum probability is also made in this work. 

\cite{Aerts13} also shares a similar opinion that quantum probability can, in fact, provide a new framework to explain several situations in which the laws of probability are violated. However, they defend that the usage of quantum probability should be more close to its real meaning in quantum mechanics and proposes to explain the phenomena, such as contextually, entanglement, observables and Flock spaces.

The work of,~\cite{Busemeyer12quantum} has stimulated some discussions on the implications of quantum probability in functional brain networks. For instance,~\cite{Stewart13} have discussed the similarities between the vector representations of quantum states with a vector symbolic architecture, which is used to model realistic biological neural models. In this sense, biological neural models can also incorporate the quantum probabilistic framework and also introduce quantum interference effects, as a mathematical and geometric alternative framework, without taking into account the true meaning that is given these effects under quantum physics. Other authors also support this idea of representing neuronal model using the quantum probabilistic framework~\cite{Banerjee13, Hameroff12, Barros09, Barros12}.

\section{Conclusions}\label{sec:conclusions}

This work was motivated by the preliminary experiments of~\cite{Tversky92} about violations of the classical probability theory on the sure thing principle. This principle states that if one chooses action $A$ over $B$ in a state of the world $X$, and if one also chooses action $A$ over $B$ under the complementary state of the world $X$, then one should always choose action $A$ over $B$, even when the state of the world in unspecified. When humans need to make decisions under risk, several heuristics are used, since humans cannot process large amounts of data. These decisions coupled with heuristics lead to violations on the law of total probability.

Recent work in cognitive psychology revealed that quantum probability theory provides another method of computing probabilities without falling into the restrictions that classical probability have in modelling cognitive systems of decision making. Quantum probability theory can also be seen as a generalisation of classical probability theory, since it also includes the classical probabilities as a special case (when the interference term is zero). 

The main difference between quantum and classical probability lies in the fact that on quantum probability we are constantly updating some beliefs when making a decision, while in classical probability all beliefs are assumed to have a definite value before a decision is made, and this value is the outcome of the decision~\cite{Aerts09}.

The main research question for this work was how could these quantum probabilities affect probabilistic graphical models, such as Bayesian Networks, since many of nowadays decision making systems are based on such structures (medical diagnosis, spam filtering, image segmentation, etc).

In this work, we proposed a novel Bayesian Network for the Computer Science community based on quantum probabilities. Our method can accommodate puzzling observations that the classical probability failed to explain (for instance, the two-step gambling game). When the nodes of the proposed Bayesian Network are represented as a superposition state, then one can look at this state as many waves moving across in different directions. These waves can crash into each other causing waves to be bigger or to cancel each other. This is the interference phenomena that the proposed Bayesian Network offers and that has direct implications when making inferences. Therefore, the proposed network represents and simulates quantum aspects motivated by Feynman's path integrals.

Experimental results revealed that the proposed quantum Bayesian Network enables many degrees of freedom in choosing the final outcome of the probabilities. If we had a real scenario, with real observations, one could use the present model to fit it to the observed data, by simply tuning the parameter $\theta$. This parameter can open a door into machine learning approaches. Learning algorithms using the proposed method might produce better prediction models, since the quantum probability amplitudes are able to fully represent real word data. For future work, we intend to explore machine learning algorithms under quantum probabilistic graphical model formalisms.

The overall results also suggested that when the classical probability of some variable is already high, then the quantum probability tends to increase it even more. When the classical probability is very low, then the proposed model tends to lower it. 

When there are many unobserved nodes in the network then the levels of uncertainty are very high. But, in the opposite scenario, when there are very few unobserved nodes, then the proposed quantum model tends to collapse into its classical counterpart, since the uncertainty levels are very low.

The proposed Bayesian Network can integrate human thoughts by representing a person's beliefs in an N-dimensional unit length quantum state vector. In the same way, the proposed quantum structure is general enough to represent any other context in which there is a need to formalise uncertainty, including prediction problems in data fusion. In the context of Bayesian Networks, data fusion is introduced in the work of~\cite{Pearl86}. The author argues that, just like people, Bayesian Networks are structures that integrate data from multiple sources of evidence and enable the generation of a coherent interpretation of that data through a reasoning process. The fusion of all these multiple data sources can be done using Bayes theorem. When a data source is unknown, then the Bayes rule is extended in order to sum out all possible values of the probability distribution representing the unknown data source. The proposed Quantum Bayesian Network takes advantage of these uncertainties by representing them in a superposition state, which enables the fusion of the data sources through quantum interference effects. These effects produce changes in the final likelihoods of the outcomes and provide a promising way to perform predictions more accurately according to reality. So, the Quantum Bayesian Network that is proposed in this work is potentially relevant and applicable in any behavioural situation in which uncertainty is involved.

\bibliographystyle{agsm}

\appendix

\section{A Grid Search Approach to Find Quantum Parameters}\label{sec:grid}

The most naive approach of parameter search is the grid search method. In this approach, it is placed a grid over the parameter space and the data is evaluated at every grid intersection, returning the parameters, which lead to the maximum performance of an algorithm \cite{Metzler07linearfeature-based}. However, grid search has the problem of being unbounded, since an infinite set of parameters are available to be tested. In the scope of this work, this is not a problem, because we always have a boundary. The values of a cosine function can all be specified for angles between the range $\left[  0, 2\pi \right]$. 

The grid search approach is a very naive method for finding parameters in the parameter space. In fact, there are several advanced parameter search algorithms in the literature, which do not have a heavy computational cost. The motivation behind the usage of the grid search approach in this work is that the computational time required by any of the other advanced methods is almost the same as using grid search. In addition, many of the advanced search parameter methods in the literature perform approximations, which can be avoided by the direct parameter search of the grid search approach \cite{Metzler07linearfeature-based}.

\begin{algorithm} [h!]
\caption{Grid Search to find $\theta$ parameters}
\label{alg:grid_search}
\begin{algorithmic}[1]
\REQUIRE classical probability when it is true ( Pr(A = t ) ), $Pr\_Ct$, \\
		~~~~~~interference term correspondent to $Pr\_Ct$, $Interference\_t$,\\
		~~~~~~classical probability when it is true ( Pr(A = f ) ), $Pr\_Cf$, \\
		~~~~~~interference term correspondent to $Pr\_Cf$,$Interference\_f$,\\
~~\\
\ENSURE Maximum quantum probabilities, \\
		~~~~~~~~~List of $\theta$ parameters that maximise the quantum probability \\
~~\\

\STATE $max\_probability \leftarrow -1$~~~~~~~~~// initialise variable that will store the maximum probability found

~~\\

\STATE $list\_parameters \leftarrow empty$~~~~~// initialise variable that will store all $theta$ parameters that maximise the probability

~~\\

\STATE // Create a for-cycle for each quantum parameter necessary to compute the probability
\FOR{$\theta_1 = 0; \theta_1 \leq 2\pi; \theta_1 = \theta_1 + 0.01 $}
\FOR{$\theta_2 = 0; \theta_2 \leq 2\pi; \theta_2 = \theta_2 + 0.01 $}

\STATE ...~~~~~~~~~~~~~~~~~~~ // Add other for-cycles to compute the other $\theta$ parameters

~~\\

\STATE Pr\_positive $\leftarrow$ Pr\_Ct + Interference\_t, $\cdot$ $\cos( \theta_1 - \theta_2 )$;
\STATE Pr\_negative $\leftarrow$ Pr\_Cf + Interference\_f, $\cdot$ $\cos( \theta_1 - \theta_2 )$;

~~\\

\STATE Pr\_positive\_norm $\leftarrow$  Pr\_positive / (  Pr\_positive +  Pr\_negative );
\STATE Pr\_negative\_norm $\leftarrow$ Pr\_negative / (  Pr\_positive +  Pr\_negative );

~~\\

 \IF{ Pr\_positive\_norm $>$ max\_probability } 
\STATE max\_probability $\leftarrow$ Pr\_positive\_norm;
\STATE list\_parameters $\leftarrow $ $\left[ \theta_1,~\theta_2 \right]$;
\ENDIF

~~\\

\ENDFOR
\ENDFOR

~~\\

\RETURN max\_probability , list\_parameters;

\end{algorithmic}
\end{algorithm}







\end{document}